%% file: arxiv-main.tex
\documentclass[sigconf]{acmart}
\usepackage{multirow}
\usepackage{algorithm}
\usepackage{algorithmic}

\AtBeginDocument{%
  }

\begin{document}

\title{A Unit Enhancement and Guidance Framework for Audio-Driven Avatar Video Generation}

% \author{Anonymous Authors}
% \settopmatter{authorsperrow=4}
\author{S.Z. Zhou}
% \orcid{}
\affiliation{
  \institution{Zhejiang University}
  \city{Hangzhou}
  \country{China}
}
% \email{}

\author{Y.B. Wang}
% \orcid{}
\affiliation{%
  \institution{Zhejiang University}
  \city{Hangzhou}
  \country{China}
}
% \email{}

\author{J.F. Wu}
% \orcid{}
\affiliation{
  \institution{Fudan University}
  \city{Shanghai}
  \country{China}
}
% \email{}

\author{T. Hu}
% \orcid{}
\affiliation{
  \institution{Shanghai Jiao Tong University}
  \city{Shanghai}
  \country{China}
}
% \email{}

\author{J.N. Zhang}
% \orcid{}
\affiliation{
  \institution{Zhejiang University}
  \city{Hangzhou}
  \country{China}
}
% \email{}

\input{secs/abstract}


% \begin{abstract}
%   A clear and well-documented \LaTeX\ document is presented as an
%   article formatted for publication by ACM in a conference proceedings
%   or journal publication. Based on the ``acmart'' document class, this
%   article presents and explains many of the common variations, as well
%   as many of the formatting elements an author may use in the
%   preparation of the documentation of their work.
% \end{abstract}


% \ccsdesc[500]{Computing methodologies~Computer vision}



\maketitle

\input{secs/introduction}

\input{secs/related_work}
\input{secs/preliminary}
\input{secs/method}

\input{secs/experiments}

\input{secs/conclusion}
\appendix
\section*{Appendix}

\section{Overview}
In this supplementary material, more details about the proposed
PAHA and more experimental results are provided, including:
\begin{itemize}
    \item More implementation details (~\autoref{sec:details});
    \item More details about CNAS (~\autoref{sec:cnas});
    \item More Ablation Study (~\autoref{sec:ablats});
    \item Limitations and future Work (~\autoref{sec:limits}).
\end{itemize}

\section{Additional Implementation Details of PAHA}\label{sec:details}
\subsection{UniVDM}
UniVDM is initialized from a pretrained video diffusion model~\cite{blattmann2023align}. During training with the PAR method, videos have a spatial resolution of 512×512, a fixed length of 32 frames, a batch size of 8, and 100k training steps. We use the AdamW optimizer~\cite{loshchilov2017decoupled} with a learning rate of 5e-5 and DDPM~\cite{ho2020denoising} as the noise scheduler with 1000 sampling steps. The confidence score threshold $\tau_j$ is set to 0.8, with a radius $r$ of 10. The hand/face weight hyperparameter $\omega_1$ is set to 10, while $\omega_2$ for the body region is set to 2.
\subsection{PAR}
\subsubsection{Training Details}
When training the backbone network $G$ using the PAR method, we use DWpose~\cite{yang2023effective} to extract the required pose sequences. 
The visual encoder from the multimodal CLIP-Huge model~\cite{sohl2015deep} in Stable Diffusion v2.1 is used to encode CLIP embeddings of reference images.

\subsection{PCE}
\subsubsection{Motivation}
\begin{figure}[htbp]
    \centering
    \includegraphics[width=.98\linewidth]{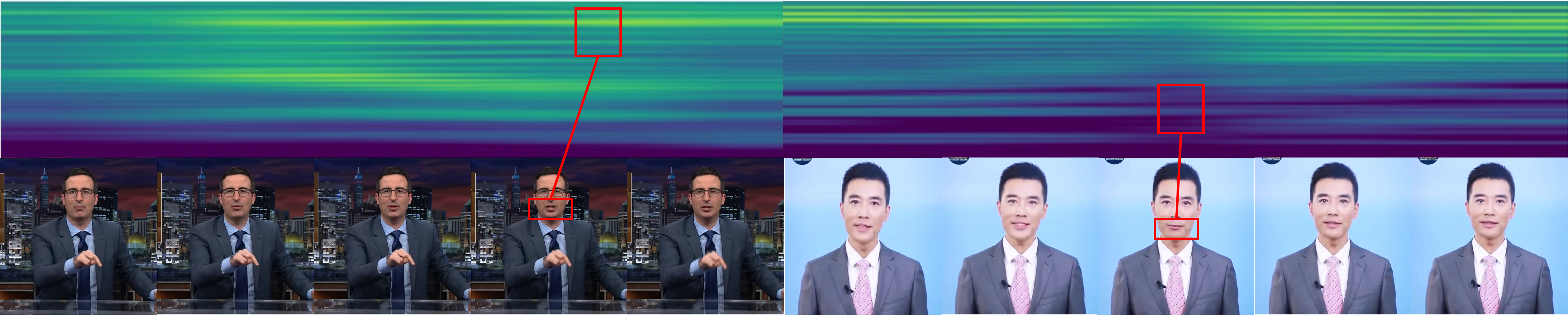}
    \caption{\label{fig:correlation}  \textbf{The correlation between motions and the corresponding spectrogram.} When the speaker begins to talk, the mid-to-high frequencies in the spectrum brighten, while darker areas of the spectrum correspond to smaller motion amplitudes.}
\end{figure}

Fig.~\ref{fig:correlation} shows the correlation between motions and the spectrogram between spectral energy changes and localized motions (hands, lips). When the speaker begins to talk, the mid-to-high frequencies in the spectrum brighten, while darker areas of the spectrum correspond to smaller motion amplitudes.

\subsubsection{Dataset Construction for Classifier}
We trained the PCE using offline training, with positive samples coming from the ground truth and negative samples generated using the pre-trained UniVDM (60k steps checkpoints).

\subsubsection{Training Details}
In the PCE classifier training experiments, all negative samples are generated using a DDIM~\cite{song2020denoising} sampler, with 30 sampling steps. We use MediaPipe~\cite{lugaresi2019mediapipe} to obtain facial boxes. Both the non-face classifier $D_{non\text{-}face}$ and face classifier $D_{face}$ use a learning rate of 1e-4, with a batch size of 16 and 100k training steps. A warmup strategy is applied for the first 5000 steps, and the encoders of both classifiers are initialized with pre-trained $G$ checkpoints. When training the classifier, we used the same noise scheduler (DDPM) as when training the Unified Video Diffusion Model (UniVDM).

\subsection{Inference}
\subsubsection{Long Video Generation}
Memory constraints make generating long videos in a single pass unfeasible. Thus, multiple short video segments must be synthesized separately and merged. Existing methods typically use a sliding window strategy to synthesize short videos from overlapping local windows and merge them by averaging the overlaps, but this can cause segment discontinuities.

In our Unified Diffusion Model, we propose a unified noise input that allows random noise videos or first-frame conditioned videos as input for video synthesis. The first-frame conditioning method uses the initial frame as the condition for generating videos starting from the frame. By utilizing this strategy, the last frame of the previous short video segment can serve as the first frame of the next segment, achieving seamless and visually coherent long animation.

\subsubsection{Algorithm}
For final inference, we use a 30-step DDIM sampler, applying classifier guidance only in the first 15 steps. The guidance strengths for the face classifier, non-face classifier, and differential guidance ($\lambda_{face}$, $\lambda_{non\text{-}face}$, $\lambda_{diff}$) are set to 0.1, 1, and 0.25, respectively.
We set $\lambda_{face}$ significantly lower than $\lambda_{non\text{-}ace}$ because experiments show that the face classifier tends to dominate the generation process more easily than the non-face classifier, and excessive $\lambda_{face}$ notably decreases video quality. 
Furthermore, we observe that both classifiers improved alignment and motion metrics but also caused varying degrees of FVD decline, with the face classifier having a more pronounced impact. Therefore, in Sequential Guidance (SG), we first use the non-face classifier and apply a smaller guidance weight to the face classifier to mitigate its influence.
The evaluation resolution of the final video is 256$\times$256.

\begin{figure}[htbp]
    \centering
    \includegraphics[width=.98\linewidth]{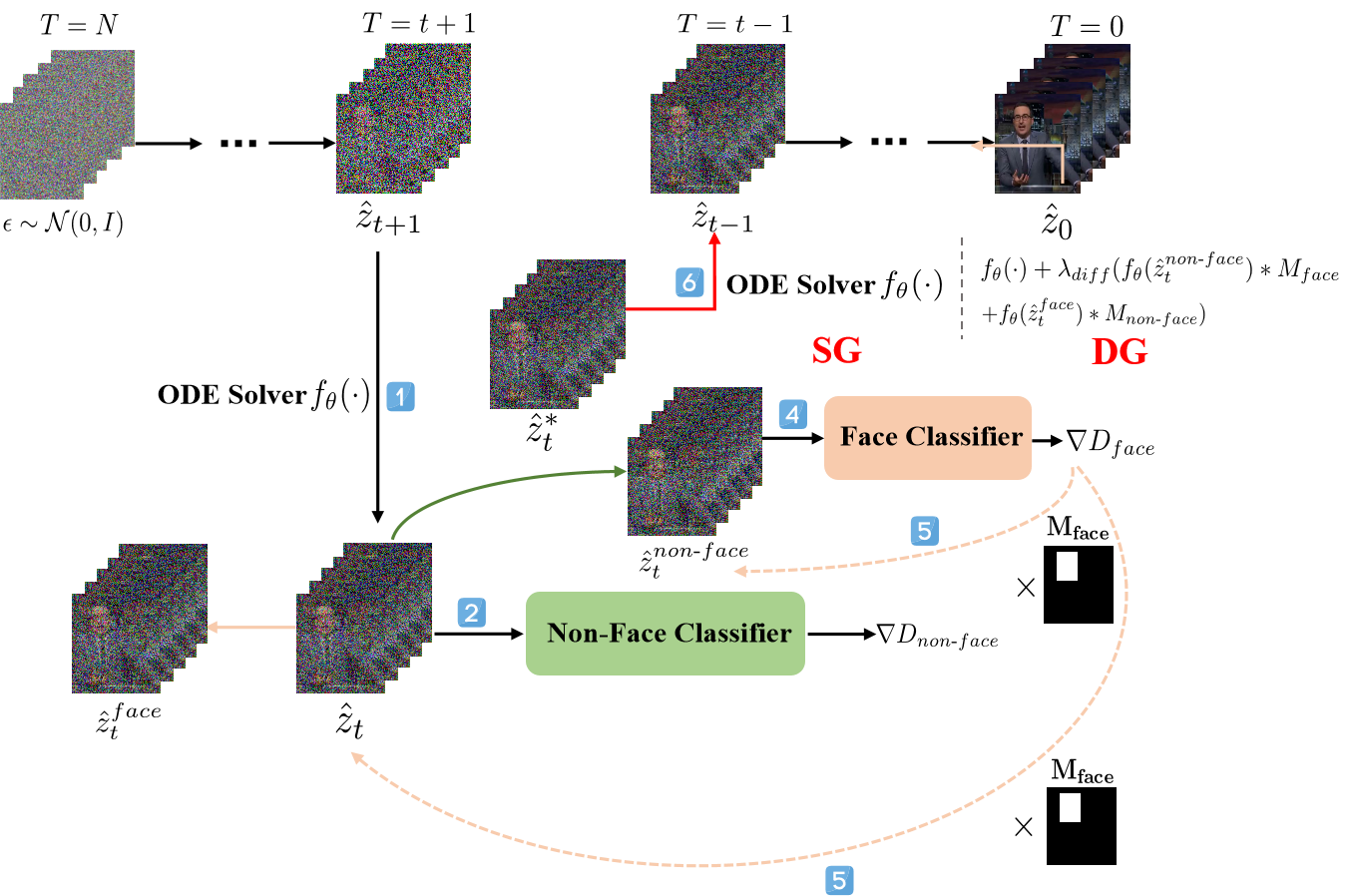}
    \caption{\label{fig:inference}  \textbf{The inference pipeline of PAHA includes two forms: PAHA-SG (\textit{Sequential Guidance}) and PAHA-DG (\textit{Differential Guidance}), focusing on generation efficiency and quality, respectively.}}
\end{figure}

\autoref{fig:inference} illustrate the inference pipelines for PAHA-SG(\textit{Sequential Guidance}) and PAHA-DG(\textit{Differential Guidance}). The only difference between the two inference methods lies in the final calculation of $\hat z_{t_{n-1}}$. In SG, gradients from the face and non\text{-}face classifiers are sequentially computed and fed back into the denoising video. In contrast, DG performs two additional ODE Solver $f_\theta(\cdot)$ executions to mitigate classifier negative impact in non\text{-}guided regions.
The pseudo code of our classifier-based inference process is shown in Algorithm~\ref{alg:inference}. 

\input{pcode/inference}

\subsection{Evaluation Metrics}
% We use MMPose~\cite{sengupta2020mm} to extract 2D human poses for motion-related metrics. 
To evaluate the quality, diversity, and alignment between gestures and speech, we employ: 1) Fréchet Gesture Distance (FGD)~\cite{yoon2020speech}, measuring the distribution gap between real and generated gestures in feature space; 2) Diversity (Div.), calculating the average feature distance between generated gestures. These metrics use an autoencoder trained on PATS and CNAS poses following the code from~\cite{liu2022learning}. Additionally, in accordance with S2G, we calculate the 3) Beat Alignment Score (BAS)~\cite{li2021learn}, measuring the average distance between speech and gesture beats. 4) Synchronization-C (Sync-C)~\cite{xu2024hallo} evaluates lip synchronization in generated videos, where a higher score reflects better alignment of lip motions with the audio.
For video evaluation, we use 5) Fréchet Video Distance (FVD)~\cite{unterthiner2018towards} to assess the overall quality of gesture videos, computed in feature space using an I3D classifier~\cite{wang2019i3d} pre-trained on Kinetics-400~\cite{kay2017kinetics}.

\section{More Details about CNAS}\label{sec:cnas}
% \input{tables/dataset}
% The PATS dataset contains approximately 84,000 clips from 25 speakers. featuring transcribed poses, aligned audio, and text transcriptions, with an average length of 10.7 seconds and a total duration of 251 hours.

With the goal of modeling human bodies, we estimate joints of 2D body and hands, and fit statistical 2D human body models by minimizing projection errors and temporal differences between consecutive frames. We further employ the advanced detection model DINO v2~\cite{oquab2023dinov2} to assist with body part detection. We filter out videos with significant screen switching, undetected or partially detected faces or bodies, unstable detection results, and poor audio quality. This process generates a dataset of 36,825 seconds with 5 identity IDs, containing 1,473 valid clips per news anchor, with a video resolution of 512$\times$896. Each clip contains 125 frames at 25 fps (5 seconds) with audio sampled at 22 kHz.

\section{More Ablation Study}\label{sec:ablats}
\noindent \textbf{Guidance Parameters. }We determine the guidance strength during inference via ablation experiments. Tables~\ref{tab:face},~\ref{tab:hand}, and~\ref{tab:diff} present the quantitative results for non-face guidance weight $\lambda_{non\text{-}face}$, face guidance weight $\lambda_{face}$, and differential guidance weight $\lambda_{diff}$ when executing Differential Guidance during the inference. The final weight combination is set to $(\lambda_{non\text{-}face}=1, \lambda_{face}=0.1, \lambda_{diff}=0.25)$, achieving optimal performance.
\input{tables/hand_weight}
\input{tables/face_weight}
\input{tables/diff_weight}

\section{Limitations and Future Work}\label{sec:limits}
Although the proposed method PAHA improves regional generation quality and motion-audio synchronization, our categorization of optimized areas remains broad, encompassing only face, hands, and body lacks finer control. This results in the model still needing improvement in realistically rendering complex expressions (emotional changes) and complex gestures (intersections and overlaps). Meanwhile, the fixed perspective of generated videos indicates the model's limitations in handling dynamic scenes. Furthermore, our Chinese co-speech dataset CNAS remains smaller than mainstream English datasets (PATS) in terms of character IDs and the diversity of actions and expressions. However, the experimental results show that although the dataset is small in scale, it is sufficient to validate the algorithm's effectiveness.

In the future, we plan to develop real-time feedback mechanisms to enhance the interactivity and realism of human animation, improving the model's robustness across viewpoints and interactions for broader applications in live media and augmented reality. Additionally, we aim to incorporate diverse speaking styles and emotions to enhance expressiveness and control, and further expand the Chinese co-speech dataset CNAS.

\bibliographystyle{Reference-Format}
\bibliography{sample-base}

\end{document}

%% file: secs/abstract.tex
\begin{abstract}
Audio-driven human animation technology is widely used in human-computer interaction, and the emergence of diffusion models has further advanced its development. Currently, most methods rely on multi-stage generation and intermediate representations, resulting in long inference time and issues with generation quality in specific foreground regions and audio-motion consistency. These shortcomings are primarily due to the lack of localized fine-grained supervised guidance. To address above challenges, we propose Parts-aware Audio-driven Human Animation, \textit{PAHA}, a unit enhancement and guidance framework for audio-driven upper-body animation. We introduce two key methods: Parts-Aware Re-weighting (PAR) and Parts Consistency Enhancement (PCE). PAR dynamically adjusts regional training loss weights based on pose confidence scores, effectively improving visual quality. PCE constructs and trains diffusion-based regional audio-visual classifiers to improve the consistency of motion and co-speech audio. Afterwards, we design two novel inference guidance methods for the foregoing classifiers, Sequential Guidance (SG) and Differential
Guidance (DG), to balance efficiency and quality respectively. Additionally, we build CNAS, the first public Chinese News Anchor Speech dataset, to advance research and validation in this field. Extensive experimental results and user studies demonstrate that PAHA significantly outperforms existing methods in audio-motion alignment and video-related evaluations. The codes and CNAS dataset will be released upon acceptance.

\end{abstract}

%% file: secs/introduction.tex
\begin{figure*}[htbp]
    \centering
    \includegraphics[width=.98\linewidth]{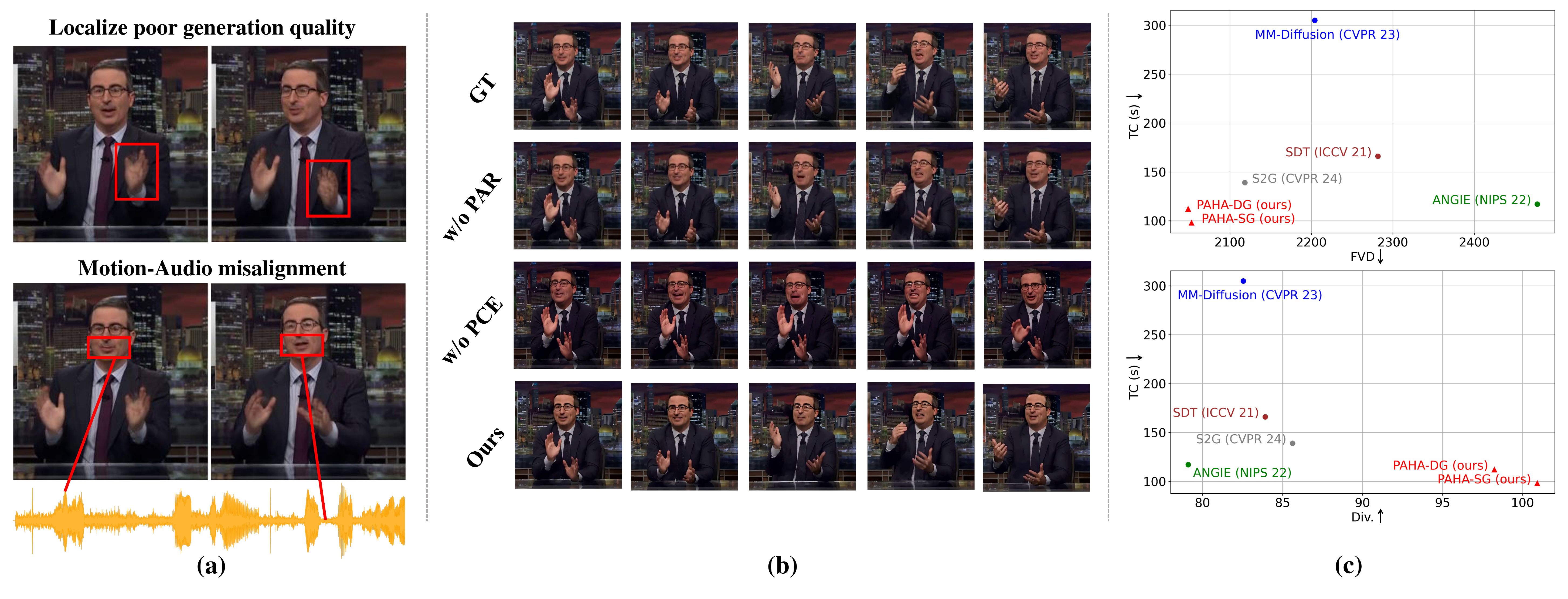}
    \caption{\label{fig:teaser} 
    \textbf{\textit{(a)}} S2G~\cite{he2024co}, the state-of-the-art for co-speech gesture generation, suffers from localized poor quality (\textit{e.g.}, hands, face) and audio-motion misalignment.
    \textbf{\textit{(b)}} Qualitative ablation study. Parts-Aware Re-weighting (PAR) improves local generation quality of characters, while Parts Consistency Enhancement (PCE) enhances alignment between motion and co-speech audio. Our comprehensive method generates high-quality and consistent videos.
    \textbf{\textit{(c)}} Comparison of baselines and our PAHA in terms of video quality (FVD, lower is better), diversity (Div., higher is better), and inference efficiency (TC\textit{(Time Cost)}, lower is better). Our method (PAHA-DG and PAHA-SG) achieves superior FVD and Div. performance while maintaining lower TC.
    }
\end{figure*}
\section{Introduction}
Human-centered content generation has been a focal point in computer vision research. Human animation technology aims to synthesize speaking character videos from a single static reference character image and corresponding speech audio. This technology holds significant value across various fields, including video games, virtual reality, film and television production, social media, digital marketing, online education, human-computer interaction, and virtual assistants~\cite{nazarieh2024survey}.

Recent advances in GAN and diffusion models~\cite{corona2024vlogger, wei2024aniportrait, tian2025emo, xu2024vasa, song2022audio, zhu2025champ, zhang2024tora, yao2024fd2talk, liu2024anitalker} have enhanced high-resolution, high-quality character generation, particularly for maintaining long-term identity consistency. For instance, Live Portrait~\cite{guo2024liveportrait} uses GANs for portrait animation with stitching and redirection controls, while diffusion-based methods like VASA-1~\cite{xu2024vasa}, EMO~\cite{tian2025emo}, and Hallo~\cite{xu2024hallo} enable end-to-end human animation. However, these methods often struggle with 
\textbf{\textit{poor generation quality in specific regions (e.g., lips, eyes) and focus primarily on audio-driven talking faces, neglecting gestures and body motions}}, 
which restricts their applicability.
Actually, co-speech gesture generation~\cite{he2024co, liu2022audio, qian2021speech}, treated as a separate task, highlights gestures' role in enhancing speech clarity, persuasion, and credibility. For example, 3D-GestureGAN~\cite{mahapatra2025co} produces realistic talking videos with natural gestures using UV texture optimization and conditional GANs, while S2G~\cite{he2024co} combines thin-plate-splines (TPS) transformations with a transformer-based diffusion model to generate long-duration, high-quality co-speech gesture videos. Despite progress, these methods face \textbf{\textit{challenges in achieving speech-motion synchronization}} due to multi-stage training and inference, leading to higher computational costs and error accumulation. The reliance on intermediate representations, such as 2D/3D landmarks, TPS, 3D mesh, and optical flow, complicates processes further. These limitations primarily stem from \textbf{\textit{the lack of localized fine-grained supervised learning or guidance}}.
% \zjn{Summarize the current challenges and present them for clear presentation using bold/textit high-light. \textbf{\textit{1)}} xxx. \textbf{\textit{2)}} xxx.}
% \zjn{better to draw a teaser figure or qualitative vis to expose these challenges}

Motivated by the divide-and-conquer strategy, we propose \textbf{P}arts-Aware \textbf{A}udio-Driven \textbf{H}uman \textbf{A}nimation (PAHA). Our method independently optimizes co-speech face and gesture generation in distinct spatial areas within a single video diffusion model, effectively addressing the limitations of existing methods regarding local poor quality and audio-video misalignment (Fig.\ref{fig:teaser}). 
Parts-Aware Re-weighting (PAR) (Sec.\ref{sec:par}) assigns dynamic loss weights to key areas (\textit{e.g.}, hands, face, body) in the training video frames. Amplifying regional loss enhances the model's focus and learning in these areas, improving generation quality.
Parts Consistency Enhancement (PCE) (Sec.\ref{sec:pce}) trains diffusion-based regional classifiers to distinguish audio-visual differences between generated and real samples, learning the temporal correlation between spectral energy variations and localized character motion. During inference, classifiers provide gradient alignment signals to guide the model's focus on specific regions, enhancing audio-video alignment. Subsequently, we design two innovative inference guidance methods for the aforementioned classifiers: Sequential Guidance (SG) (Sec.\ref{sec:SG}) for efficiency and Differential Guidance (DG) (Sec.\ref{sec:DG}) for quality.
% This dataset comprises videos featuring dynamic gestures, which highlights the complexities of human communication. Datasets in various languages are crucial for thoroughly evaluating the performance of different methods. We validate its effectiveness against our method and baselines.

Additionally, given the lack of open-source Chinese co-speech gesture datasets, we constructed the Chinese News Anchor Speech Dataset (CNAS). This dataset comprises videos featuring dynamic gestures, which highlights the complexities of human communication. Multilingual datasets are crucial for evaluating methods comprehensively. We validate its effectiveness against our method and baselines.

Our contributions are summarized as follows:

\begin{itemize}
    \item We propose PAHA, an end-to-end framework for generating audio-driven upper-body human animation. Parts-Aware Re-weighting (PAR) method dynamically adjusts loss based on keypoint confidence scores, improving animation quality.
    \item We introduce Parts Consistency Enhancement (PCE), which uses self-distillation to train diffusion-based regional classifiers, enabling them to learn the temporal correlation between character motion and audio spectrum.
    \item During inference, we apply classifier-based consistency guidance to align regional motion with audio, offering two approaches: Sequential Guidance (SG) for efficiency and Differential Guidance (DG) for quality.
    % Additionally, PCE supports plug-and-play functionality and is easily extendable to other methods. 
    \item We create the Chinese News Anchor Speech Dataset (CNAS) to address the lack of Chinese co-speech datasets and validate multiple methods on it.
    \item Extensive experiments show that our framework produces high-quality, lifelike animations aligned with audio and outperforms previous state-of-the-art methods in quantitative and qualitative evaluations.
\end{itemize}

%% file: secs/related_work.tex
\section{Related Work}
\subsection{Audio-Driven Human Animation}
% The progress of audio-driven human animation has greatly benefited from advancements in audio understanding and driving capabilities. 
Audio-driven human animation methods can be divided into two categories: talking head generation and co-speech gesture generation. Talking head generation focuses on head motion and facial expression quality. 
% For example, SadTalker~\cite{zhang2023sadtalker} and and HEAD~\cite{liu2024audio} have effectively addressed challenges related to facial synchronization and expression modulation, achieving dynamic lip movements and coherent head motions. 
These methods commonly use 3D meshes, 2D/3D landmarks, NeRF, segmentation, or optical flow to enhance control over head movement, gaze, and blinking~\cite{xu2024vasa, xiong2024segtalker, zhou2021pose, yang2024consistentavatar, chatziagapi2024talkinnerf}. 
% Studies such as DreamTalk~\cite{ma2023dreamtalk} and EMO~\cite{tian2025emo} emphasize the importance of emotional expression and attempt to integrate audio cues with facial dynamics. 
For instance, VASA-1~\cite{xu2024vasa} utilizes 3D-aided representations in facial latent space to decouple features and generate high-quality faces. % Vividtalk~\cite{sun2023vividtalk} separately generates head movements and facial expressions using a 3D facial mesh as an intermediate representation.
% FD2Talk~\cite{yao2024fd2talk} decouples facial details into motion and appearance. In the initial phase, it predicts motion coefficients from audio using a diffusion transformer, captures appearance textures by encoding reference images, and finally uses these features as conditions to guide frame generation.
ConsistentAvatar~\cite{yang2024consistentavatar} introduces a time-sensitive detail (TSD) map with a temporally diffusion module to align results with video frames. 
SegTalker~\cite{xiong2024segtalker} decouples semantic regions into style codes, enabling talking segmentation driven by speech.
% AniTalker~\cite{liu2024anitalker} uses self-supervised learning to reconstruct target video frames from source images, learning subtle motion information and minimizing identity features through an identity encoder to achieve identity decoupling.
PersonaTalk~\cite{zhang2024personatalk} uses cross-attention to inject speaking style into audio features, ensuring lip-sync accuracy. TalkinNeRF~\cite{chatziagapi2024talkinnerf} integrates body posture, gestures, and facial expressions in a unified dynamic NeRF to generate animations with detailed hand and facial movements. However, these methods often suffer from limited intermediate representation capacity, which constrains video realism.
Co-speech gesture generation further extends to the movement of gestures.
ISCG~\cite{ginosar2019learning} generates 2D skeletal gestures from audio and synthesizes them through a pose-to-image network. SDT~\cite{qian2021speech} uses gesture template vectors combined with audio to create natural, synchronized upper body movements. ANGIE~\cite{liu2022audio} enhances 2D skeletal gestures by integrating learned template vectors and a gesture codebook, modeling body movements with unsupervised MRAA features, but often produces unnatural and inaccurate gestures. DiffTED~\cite{hogue2024diffted} improves temporal consistency and diversity in gestures using a thin-plate spline (TPS) motion model. Make-Your-Anchor~\cite{huang2024make} introduces shape constraints with 3D mesh conditioning and diffusion models, enabling precise torso and hand movements in anchor-style videos.
Nevertheless, these methods still struggle with the generation quality of hand and lip regions and rely on multi-stage generation.
In contrast, our method enables end-to-end generation without relying on other representations and focuses on regional supervised learning, enhancing the quality of specified parts.

\subsection{Diffusion-based Video Generation}
The diffusion-based model VDM~\cite{ho2022video} extends U-Net~\cite{ronneberger2015u} to capture temporal information for video generation. Make-A-Video~\cite{singer2022make} leverages pretrained text-to-image (T2I) models for efficient training without requiring paired text-video data. AnimateDiff~\cite{guo2023animatediff} integrates a motion module for seamless use with T2I models, enabling temporally coherent animations. Beyond text-based methods, additional conditions like pose, skeletons, and audio enhance control. Animate Anyone~\cite{hu2024animate} and DisCo~\cite{wang2023disco} use pose conditioners for motion guidance, while Champ~\cite{zhu2025champ} employs SMPL for unified body representation with 2D skeleton guidance. MM-Diffusion~\cite{ruan2023mm} achieves joint audio-video generation using dual U-Nets for aligned outputs. Our method operates in latent space, ensuring audio-video alignment with reduced inference costs.

%% file: secs/preliminary.tex
\section{Preliminary}\label{sec:pre}

Video Diffusion Models (VDM)~\cite{chen2024videocrafter2, guo2023animatediff, wang2023modelscope} extend image diffusion models for video generation by learning video distributions via denoising samples from a Gaussian distribution. A learnable autoencoder compresses videos into latent representations $z = E(x)$, and the diffusion model $\epsilon_\theta$ predicts noise $\epsilon$ at time $t$, conditioned on text $c_{text}$. The training objective is:
\begin{equation}\label{eq:diffusion-loss}
    L_{video}=\mathbb E_{z,c,\epsilon \sim \mathcal N(0,I),t}\|\epsilon-\epsilon_\theta(z_t,c_{text},t)\|^2,
\end{equation}
Here, $z \in R^{F\times H \times W \times C}$ denotes the latent video code. $z_t$ is derived as $z_t = \lambda_t z_0 + \sigma_t \epsilon$, with $\sigma_t = \sqrt{1-\lambda_t^2}$ controlling the diffusion scheduler. Studies~\cite{peng2024controlnext} add control signals like image $c_{img}$ or audio $c_a$. During inference, the model denoises $z_T \sim \mathcal N(0, I)$ to $z_0$, and the frozen decoder reconstructs the video.

%% file: secs/method.tex
\section{Method}

\begin{figure*}[tbp]
    \centering
    \includegraphics[width=.98\linewidth]{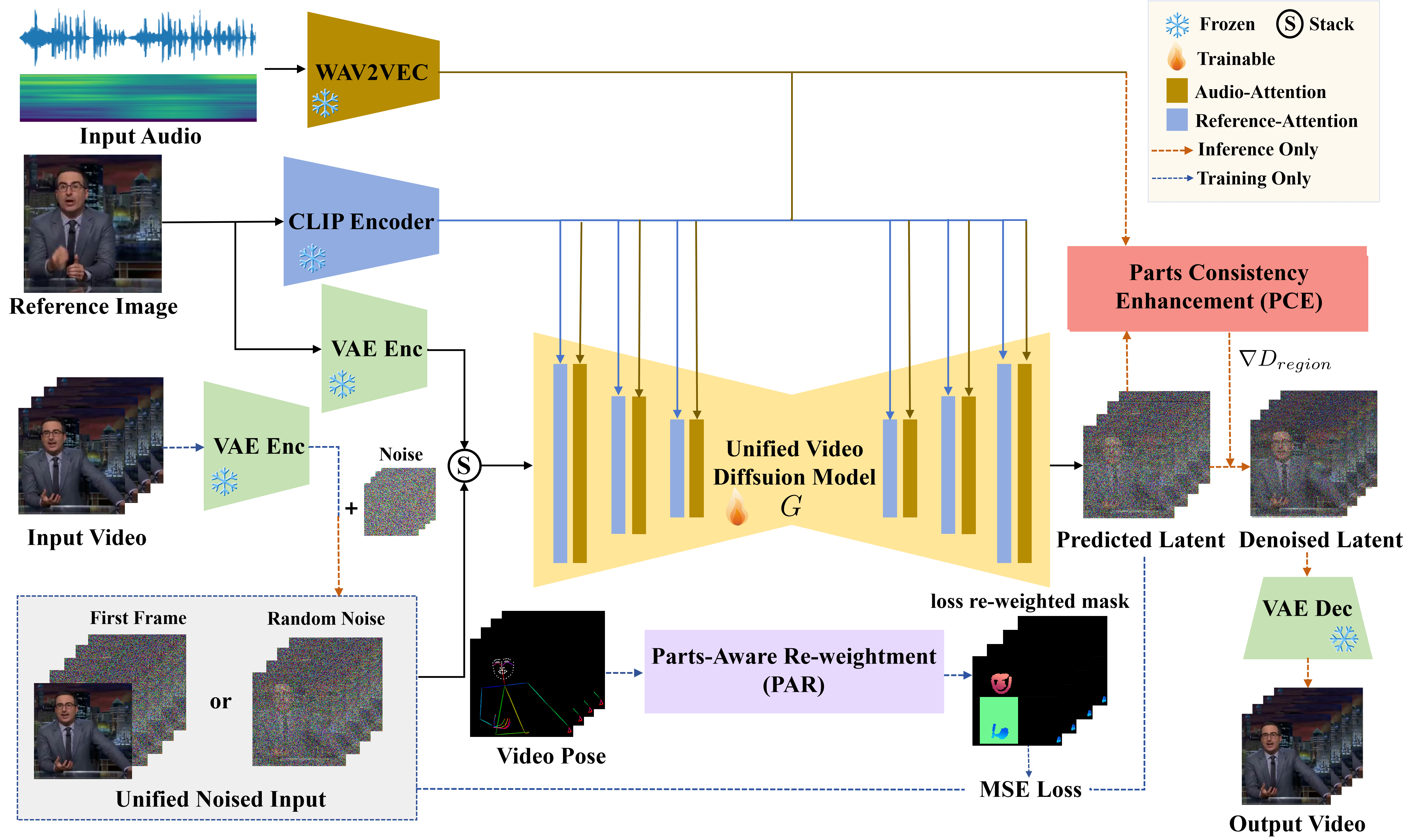}
    \caption{\label{fig:pipeline} 
    \textbf{Overview of the proposed PAHA} that consists of three core components: 
    \textbf{\textit{(a)}} The backbone of the Unified Video Diffusion Model (UniVDM) is a 3D U-Net. Video frames are encoded using a VAE encoder, while latent features of the reference image are extracted with both the CLIP and VAE encoders. These features are concatenated with the noisy input along the channel dimension, derived from either a conditioned first frame video or a noise video. 
    % Audio embeddings, processed by a pre-trained audio encoder, are fed into the 3D U-Net backbone, which uses reference and audio attention to denoise the input.
    \textbf{\textit{(b)}} Parts-Aware Re-weighting (PAR) generates a dynamic loss re-weighting mask from confidence scores of video pose keypoints, improving supervision in specific regions during training.
    \textbf{\textit{(c)}} Parts Consistency Enhancement (PCE) operates during inference, generating consistency gradients from audio and noise video features to enhance temporal visual-audio consistency.}
\end{figure*}

This section introduces Parts-Aware Audio-Driven Human Animation (PAHA), an end-to-end audio-driven framework for generating half-body human animations with a diffusion model. Fig.~\ref{fig:pipeline} illustrates the overview of our PAHA. Given a speech audio $a$ and a reference character image $I_\mathrm{ref}$, the framework generates a motion video aligned with the audio.

The process is formulated as: $V=G_\mathrm{PAR}(I_\mathrm{ref},a,\nabla_\mathrm{PCE}(z_t,a))$, where $G_\mathrm{PAR}(\cdot)$ denotes the unified video diffusion model trained with the PAR method, which simultaneously handles reference images and noisy videos without reference network. 
PAR identifies ``Awareness Areas'' we defined in key regions of the character (\textit{e.g.}, hands, face, body) in the training video frames based on a pose confidence-aware score, then apply dynamic loss weights in these areas to improve temporal smoothness and generation quality.
$\nabla_\mathrm{PCE}(z_t,a)$ represents the alignment guidance gradient produced during inference by diffusion-based regional classifiers trained using the PCE method. This gradient, based on noised latent features and audio conditions, ensures gestures and facial motions are consistent with the audio. The core process of PCE is the construction of classifiers and preparation of training samples. For inference, we design two guidance methods: Sequential Guidance for efficiency and Differential Guidance for quality. The upcoming sections cover the Unified Diffusion Model (Section~\ref{sec:pipeline}), PAR (Section~\ref{sec:par}), PCE (Section~\ref{sec:pce}), and the Inference Method (Section~\ref{sec:inference}). Section~\ref{sec:dataset} introduces the Chinese News Anchor Speech Dataset (CNAS), a co-speech Chinese dataset we created.
\subsection{Parts-Aware Audio-Driven Animation}
\subsubsection{\textbf{Unified Video Diffusion Model (UniVDM)}}\label{sec:pipeline} 

We construct the diffusion model backbone network $G$ for video generation, as shown in Fig.\ref{fig:pipeline}. To ensure temporally consistent character animation, we use the widely adopted 3D-UNet structure\cite{blattmann2023align, wang2024videocomposer}. $G$ would denoise multi-frame noisy latent inputs into continuous video frames at each timestep, conditioned on a reference character image and driving audio.

\noindent \textbf{Backbone Network. } Diffusion-based video generation frameworks often use ControlNet-like 3D-UNet models~\cite{zhang2023adding, hu2024animate, guo2023animatediff} to maintain temporal coherence, incorporating a reference encoder that replicates the 3D-UNet without temporal transformer layers to preserve the reference image's details. However, these methods typically rely on multiple large networks, increasing parameter counts and optimization challenges.

To address this, we propose the unified video diffusion model (UniVDM), which processes reference images and noisy videos simultaneously by embedding reference information and estimating video content within a shared feature space. This design aligns features and ensures temporally coherent generation without requiring an additional reference encoder, reducing model parameters. The reference image is first encoded into latent space using a VAE encoder, producing a feature representation $f_{ref}$ with dimensions $C\times h \times w$ (channels, width, height). Next, the reference representation $f_{ref}$ and video features $f_v$ are stacked along the temporal dimension to form a merged feature $f_{merge} \in R^{(t+1) \times C \times h \times w}$, where $t$ is the temporal length. Finally, these combined features are processed by the unified diffusion model.

\noindent \textbf{Audio layer. } Audio is the primary signal driving the diffusion model $G$ to generate character animations. We create audio representation embeddings $A(f)$ for each frame by concatenating features from various modules of the pre-trained wav2vec~\cite{schneider2019wav2vec}. Since motions can be influenced by future or past audio segments, such as mouth openings or inhalations before speaking, we define the speech features for each frame as:
$A(f) = \oplus\{A(f-m), ...A(f), ...A(f+m)\}$, 
where $m$ represents the number of additional features on each side. To incorporate speech features into the generation process, we add audio attention layers after each reference attention layer in the backbone network, performing cross-attention between latent features $z_t$ and $A$: $z_t = \mathrm{CrossAttention}(z_t, A(f))$. 
\subsubsection{ \textbf{Parts-Aware Re-weighting (PAR)}}\label{sec:par} 
Experimental results show that while UniVDM effectively generates audio-driven half-body character animations, the quality of hand and face regions remains suboptimal. This suggests that relying solely on the pre-trained diffusion video model and cross-attention modules, $G$ faces challenges in fine-grained supervised learning for specific areas.

\noindent \textbf{Motivation. }
The loss function of the diffusion model (Eq.\ref{eq:diffusion-loss}) assumes a uniform spatial prior; however, video frames often exhibit imbalances across foreground regions. Sequential frames can create uncertainties in dynamic appearance and motion, which negatively impact pose estimation and affect both training and inference. Furthermore, noisy pose guidance may lead to overfitting on misaligned samples, causing training instability. 

\begin{figure}[tbp]
    \centering  
    \includegraphics[width=\linewidth]{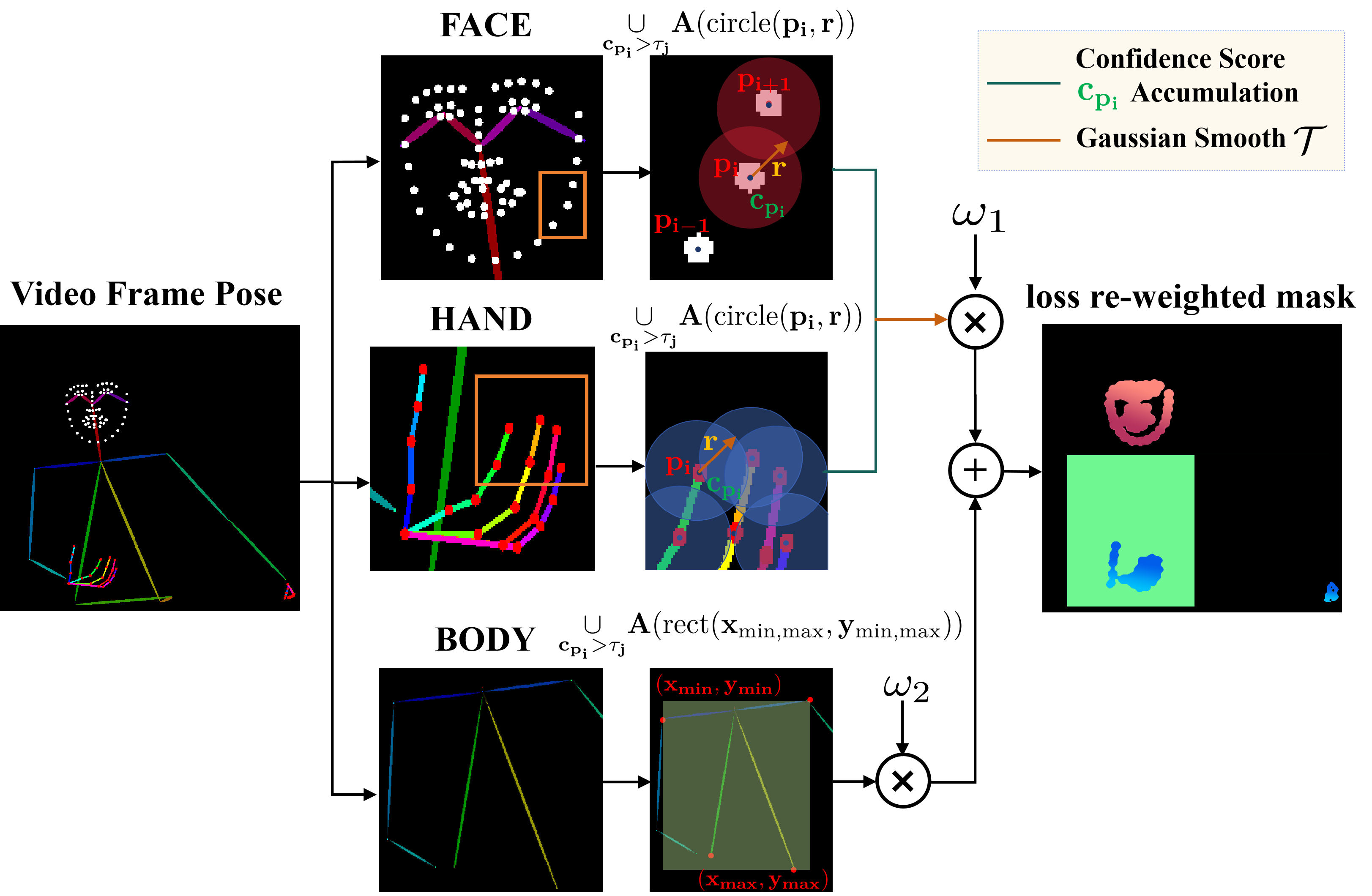}  
    \caption{\textbf{Process of our Parts-Aware Re-weighting (PAR) method.} We identify the ``Awareness Area'' for the hand, face, and body regions based on pose keypoints and their confidence scores, then independently calculate weighted confidence scores for each region and merge them into a loss re-weighted mask.}
    \label{fig:par}  
\end{figure}

To tackle these issues, we propose the Parts-Aware Re-weighting (PAR) method ( shown in Fig.~\ref{fig:par}), which allows for dynamic loss re-weighting of specific regions to improve foreground generation.
%Actually, the generalized objective loss function of the diffusion model (Eq.\ref{eq:diffusion-loss}) assumes a uniform prior distribution in spatial coordinates, but there is a significant imbalance between different foreground regions in video frames. Additionally, sequential frames often introduce uncertainties in dynamic appearance and motion compared to single 2D training images, leading to inaccurate pose estimation that impacts training and inference. Furthermore, noisy pose guidance signals can result in overfitting on incorrectly posed samples, potentially causing training instability.

% To address this issue, we propose the Parts-Aware Re-weighting (PAR) method to achieve dynamic loss re-weighting for arbitrary regions, helping the model better focus on foreground generation within specific areas. The overall process is shown in Fig.~\ref{fig:par}
%``Awareness Area''.
\noindent \textbf{Awareness Area. }Following the approach of ConvoFusion~\cite{mughal2024convofusion}, we divide the half-body character into three regions: hand, face, and body. Specifically, PAR leverages the confidence scores associated with each keypoint in the pose estimation model, where higher scores reflect better visual quality (less blur and occlusion). We establish a confidence score threshold $\tau_j$, considering keypoint $p_i$ with score $c_{p_i}$ above it as reliable. Then we generate loss re-weighting masks based on these thresholds. 
% For hand and face regions, we draw circles with radius $r$ around reliable keypoints and take their union. For the body region, we create a rectangle using the extreme x and y coordinates of all body keypoints. Thus, the ``Awareness Area" $S_\mathrm{a}$ is defined as:
For pixels $x$ in smaller, dynamic regions (hand and face), circles with radius $r$ are drawn around reliable keypoints and merged for precise delineation. For the stable larger region (body), a rectangle is formed using the extreme x and y coordinates of all body keypoints. The ``Awareness Area'' $S_\mathrm{a}$ is thus defined as:
% \begin{align}
%     S_\mathrm{a} &= \left\{  
%     \begin{array}{ll}  
%         \underset{c_{p_i}>\tau_j} \cup A(\text{circle}(p_i,r)) & \text{if } p_i \in S_\text{hand} \cup S_\text{face} \\ %[1ex]  
%         \underset{c_{p_i}>\tau_j} \cup A(\text{rect}(x_{\text{min,max}}, y_{\text{min,max}})) & \text{if } p_i \in S_\text{body} \\  
%     \end{array}  
%     \right.
% \end{align}
\begin{equation}
\resizebox{0.91\linewidth}{!}{$
\begin{aligned}
    S_\mathrm{a} &= \left\{  
    \begin{array}{ll}  
        \underset{c_{p_i}>\tau_j} \cup A(\text{circle}(p_i,r)), & \text{if } p_i \in S_\text{hand} \cup S_\text{face} \\  
        \underset{c_{p_i}>\tau_j} \cup A(\text{rect}(x_{\text{min,max}}, y_{\text{min,max}})), & \text{if } p_i \in S_\text{body}. \\  
    \end{array}  
    \right.
\end{aligned}
$}
\end{equation}
Here, $A(\cdot)$ represents the area of the corresponding region, $\text{circle}(p_i, r)$ denotes a circle with center $p_i$ and radius $r$. $\text{rect}(x_{\text{min,max}}, y_{\text{min,max}})$ indicates a rectangle defined by the left and right x-coordinates $x_{\text{min,max}}$ and the top and bottom y-coordinates $y_{\text{min,max}}$.

%``Loss Re-weighting''

\noindent \textbf{Loss Re-weighting. }During the video diffusion model's loss computation, the ``Awareness Area'' $S_\mathrm{a}$ is assigned higher weights to prioritize its influence during training. For pixels $x$ in hand and face Awareness Areas, the mask weight is calculated by applying Gaussian smoothing to the sum of confidence scores $c_{p_i}$ of keypoints at circle centers, multiplied by the hyperparameter $\omega_1$. For body and other pixels $x$, the mask weight is set to the hyperparameter $\omega_2$, which defaults to 1 outside the body Awareness Area. The loss re-weighting mask $M(x)$ can be expressed as:
% \begin{align*}  
%     M(x) &=   
%     \begin{cases}  
%         \mathrm{Gau}\left( \sum_{p_i \in S_\text{hand} \cup S_\text{face}} c_{p_i} \right) \cdot \omega_1 & \text{if } x \in S_\text{a}^\text{hand} \cup S_\text{a}^\text{face} \\[1ex]  
%         \omega_2 \; \text{or} \; 1 & \text{if } x \in S_\text{a}^\text{body} \;\text{ else}   
%     \end{cases}  
% \end{align*}
% \begin{align}  
%     M(x) &=   
%     \begin{cases}  
%         \mathcal{T}\left( \sum_{p_i \in S_\text{hand} \cup S_\text{face}} c_{p_i} \right) \cdot \omega_1 & \text{if } x \in S_\text{a}^\text{hand} \cup S_\text{a}^\text{face} \\[1ex]  
%         \omega_2 & \text{if } x \in S_\text{a}^\text{body} \; \text{or else}  
%     \end{cases}  
% \end{align}
\begin{equation}
\resizebox{0.91\linewidth}{!}{$
\begin{aligned}  
    M(x) &=   
    \begin{cases}  
        \mathcal{T}\left( \sum_{p_i \in S_\text{hand} \cup S_\text{face}} c_{p_i} \right) \cdot \omega_1, & \text{if } x \in S_\text{a}^\text{hand} \cup S_\text{a}^\text{face} \\[1ex]  
        \omega_2, & \text{if } x \in S_\text{a}^\text{body} \; \text{or else}. 
    \end{cases}  
\end{aligned}
$}
\end{equation}
% Here, $\mathcal{T}(\cdot)$ denotes the Gaussian smoothing operation. Inspired by findings from~\cite{wang2024instancediffusion} showing inferior generation quality for small objects, we set the value of $\omega_2$ to be less than $\omega_1$ to facilitate the generation of content in small areas. 
% In summary, the PAR method directs model training toward specific regions, particularly the hand and face, improving visual quality and enhancing the realism of generated content.
Here, $\mathcal{T}(\cdot)$ represents the Gaussian smoothing operation. Following ~\cite{wang2024instancediffusion}, which highlights poorer generation quality for small objects, $\omega_2$ is set lower than $\omega_1$ to better handle smaller areas. In summary, the PAR method focuses model training on specific regions, particularly the hands and face, enhancing visual quality and realism in generated content.
\subsubsection{\textbf{Parts Consistency Enhancement (PCE)}}\label{sec:pce}
Ensuring character motion aligns with driven audio is crucial for assessing framework performance. Most existing audio-visual strategies struggle to synchronize discrete character motion with the continuous audio spectrum, causing inherent video inconsistencies.
 
\noindent \textbf{Motivation. }A speaker's body motions naturally coordinate with their spoken content, exhibiting rhythmic temporal correlations. However, we find that relying solely on single frames and long-distance audio for alignment can overlook temporal consistencies in visual features. 

Thus, We propose the Parts Consistency Enhancement (PCE) method, which trains diffusion-based regional classifiers to discriminate between real and generated videos, implicitly learning local audio-video consistency. During inference, the classifier guides the process with gradients to synchronize key movements with speech.
\begin{figure}[htbp]
    \centering  
    \includegraphics[width=0.9\linewidth]{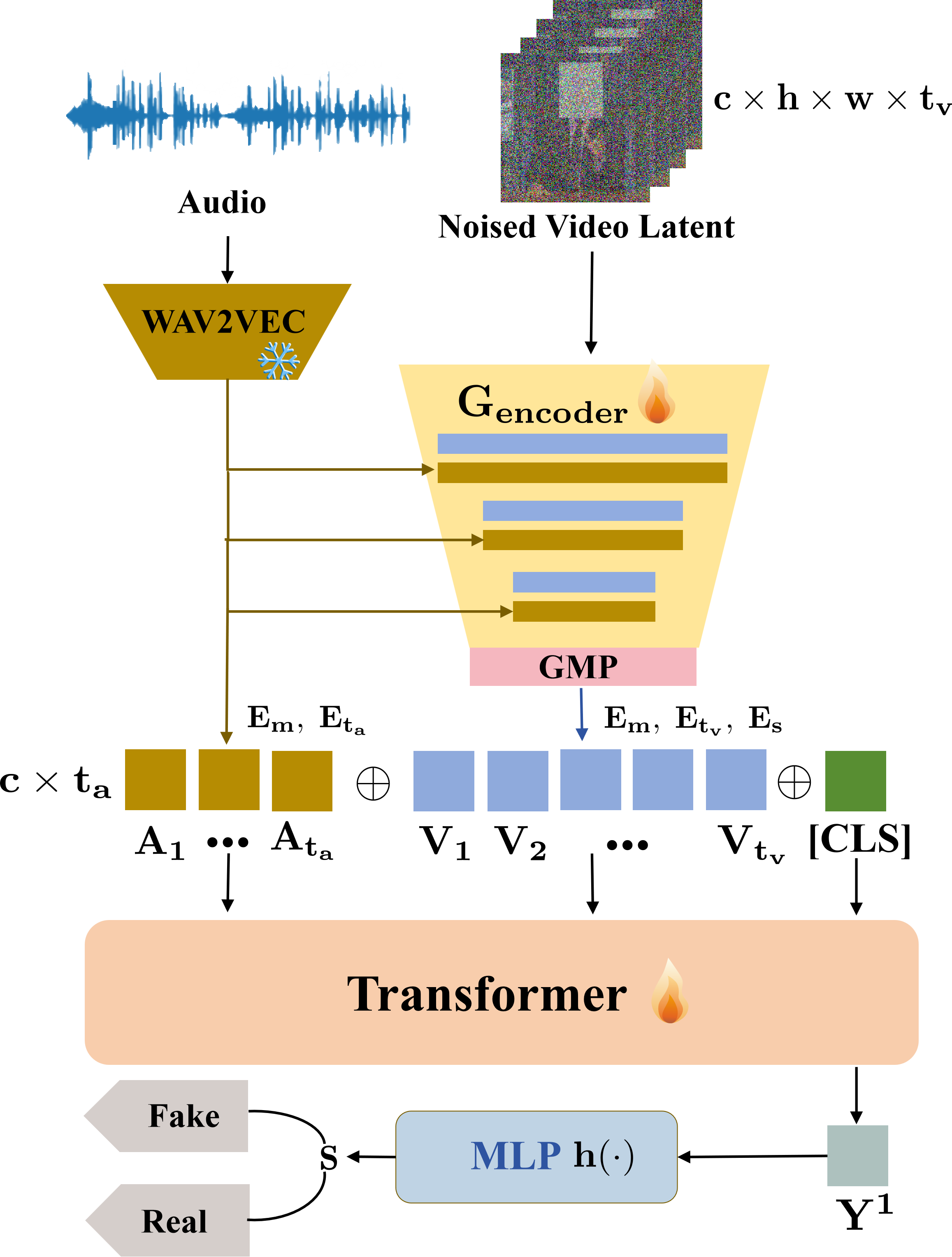}  
    \caption{\textbf{Structure of our diffusion-based classifier.} The pretrained diffusion encoder supports inputting noised video features and clean audio. After dimensionality reduction by GMP, the audio-video sequence is fed into the Transformer for full self-attention interaction, and the MLP head finally predicts the audio-video synchronization score.}
    \label{fig:classifier}  
\end{figure}

\noindent \textbf{Classifier Construction. }After training UniVDM, We start to train audio-visual classifiers combining a diffusion model U-Net encoder and a transformer encoder, as shown in Fig.~\ref{fig:classifier}. The input consists of noised video latent features $V_i$ and clean audio samples $A_j$. Prior works~\cite{xiaotackling, sauer2025adversarial, lin2024sdxl} have used adversarial discriminators in diffusion model distillation to accelerate sampling. We choose the U-Net encoder from the pre-trained Unified Diffusion Model (Section~\ref{sec:pipeline}) as the classifier's initial module, offering two advantages: 1) leveraging the diffusion model's understanding of audio-video modalities to simplify training; 2) processing latent features across all diffusion timesteps, avoiding pixel-space mapping and reducing computational costs.

Building on the above discovery regarding frequency domain and motion relationships, we need to extract time-domain features. Drawing inspiration from natural language translation tasks, we consider the correlation between motions and spectral information to be analogous to the relationship between ``vocabulary'' and ``sentence''. Latent video features $f_v \in \mathbb R^{c\times h \times w \times t_v}$ are concatenated with clean audio features $f_a \in \mathbb R^{c\times t_a}$ and serialized them into the transformer~\cite{vaswani2017attention} encoder for interaction through self-attention.

Directly flattening all visual features and densely combining them with audio features for the transformer is computationally expensive, with a quadratic complexity of $\mathcal O((hwt_v+t_a)^2)$, limiting scalability for longer videos. To address this, we apply global max pooling (GMP) to each video frame instead of dense visual inputs. We also introduce a learnable class token ($\text{[CLS]}$) to help the classifier distinguish modalities while preserving spatial-temporal positional information. We add modal encoding $E_m \in \mathbb R^{c\times 2}$ to the video and audio features, indicating the feature type (i.e., audio or visual), along with temporal encoding $E_{t_{\{v,a\}}} \in \mathbb R^{c\times t_{\{v,a\}}}$. The video features also include 3D RoPE~\cite{su2024roformer} positional encoding embeddings $ E_s \in \mathbb{R}^{c \times h \times w} $, which are efficient for varying video token counts. This process is expressed as:
\begin{align}  
    \overline V_i &= \text{GMP}(V_i) + E_m + E_{t_v} + E_s, \\
    \overline A_j &= A_j + E_m + E_{t_a}, \\
    Z_{ij} &= [\text{[CLS]} \oplus \overline V_i \oplus \overline A_j]. 
\end{align}
Here, $\oplus$ indicates a concatenation operation. The length of the input sequence $Z_{ij}$ to the transformer encoder is reduced from $(hwt_v + t_a + 1)$ to $(t_v + t_a + 1)$, resulting in significant memory savings. A multi-layer perception~\cite{taud2018multilayer} serves as the classifier $h$.

Ensuring character motion aligns with driven audio is as crucial as video quality for evaluating framework performance. Many existing audio-visual methods fail to synchronize discrete character motions with continuous audio, causing video inconsistencies.

\begin{figure}[tbp]
    \centering
    \includegraphics[width=0.9\linewidth]{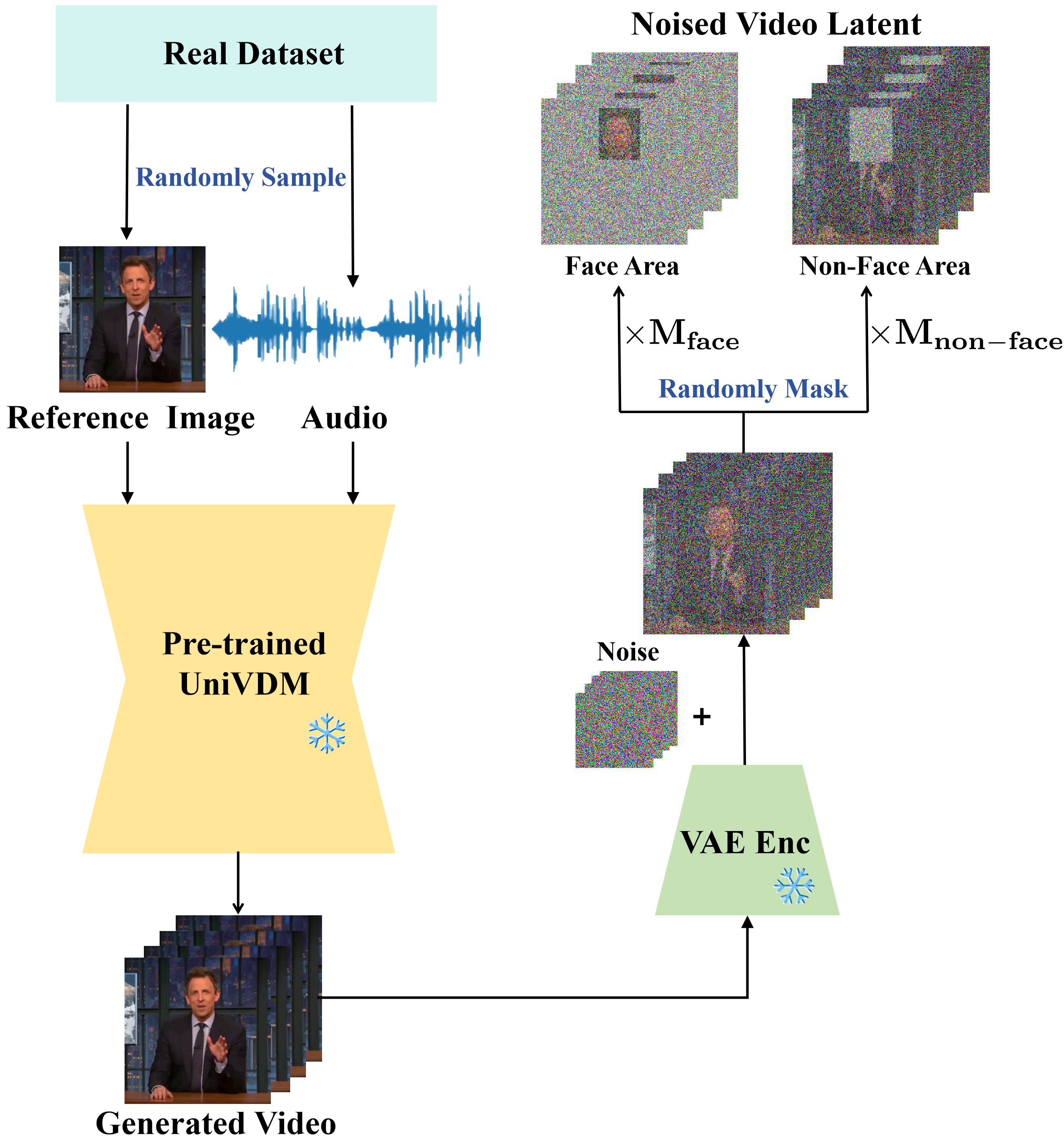}  
    \caption{\textbf{Pipeline for constructing negative samples for the classifier.} The video is generated by our pre-trained UniVDM, conditioned on randomly sampled audio and reference frames from the real dataset. We mask specific areas with a certain probability to enhance local modality alignment.} 
    \label{fig:sample}  
\end{figure}

\noindent \textbf{Training Data. }Inspired by Diffusion Self-Distillation~\cite{cai2024diffusion}, we generate negative samples using the pre-trained unified video diffusion model $G$, while positive samples come from real videos, as shown in Fig.~\ref{fig:sample}. We employ a masking strategy to help the classifier focus on consistent features between motion video and co-speech audio. We train two classifiers: \textit{non-face classifier} and \textit{face classifier}. To train the \textit{non-face(i.e., areas other than the face) classifier}, facial regions are randomly masked to minimize their influence on audio matching, emphasizing non-face motions. For the \textit{face classifier}, only facial regions are retained while masking the rest. Additionally, we use two simple data augmentation strategies: 1) random reference frame sampling during inference, starting with the first frame, and 2) varying audio and video lengths. These strategies improve the one-to-many mapping between audio and motion.

\noindent \textbf{Training Loss. }
We use only the first token output from the final encoder layer ($Y^1_{ij}$), corresponding to the $\text{[CLS]}$ position, as the aggregated representation of the entire output sequence for the MLP. The output is a synchronisation score $s_{ij}$, indicating to what degree the inputs $V_i$ and $A_i$ are in sync, $s_{ij} = h(Y^1_{ij})$. We optimize the classifier using binary cross-entropy loss:
\begin{equation}
       L_{cls}  = - (y \log(s) + (1 - y) \log(1 - s)). 
\end{equation}
Here, $y$ represents the final predicted label.

\subsection{Inference Process}~\label{sec:inference} 
The face and non-face classifier are trained separately offline but could work simultaneously during inference, each focusing on its assigned area. We propose two effective inference methods.

%\noindent (1) \textbf{Sequential Guidance}
\subsubsection{\textbf{Sequential Guidance (SG)}}~\label{sec:SG}
Start by sampling $z_{T} \sim  \mathcal{N}(0, I)$ from a Gaussian distribution. During inference, the numerical solver $f_\theta$ first predicts $\hat z_{t} = f_\theta(\hat z_{t+1}, a,t+1)$. Then the non-face classifier $D_{non\text{-}face}$ computes the gradient $\nabla D_{non\text{-}face}(\hat z_{t}, a) \in \mathbb R^{c\times h \times w}$ conditioned on $(\hat z_{t}, a)$. This gradient spatially influences only the potential features guided by the non-face mask area $M_{non\text{-}face}$, resulting in:
\begin{equation}
\resizebox{0.91\linewidth}{!}{$
    \hat z_{t}^{non\text{-}face} = \hat z_{t} - \lambda_{non\text{-}face} \sigma_t \nabla D_{non\text{-}face}(\hat z_{t}, a) * M_{non\text{-}face} ,
    $}
\end{equation}
where $\lambda_{non\text{-}face}$ is the gradient weight used to control the strength of the condition. Next, $(\hat z_{t}^{non\text{-}face},a)$ is used as the condition for the facial classifier $D_{face}$ to compute the facial consistency gradient $\nabla D_{face}(\hat z^{non\text{-}face}_{t}, a) \in \mathbb R^{c\times h \times w}$. Similarly, by introducing $M_{face}$, this gradient works only on the facial region:
\begin{equation}
   \hat z_{t}^{face} = \hat z_{t} - \lambda_{face} \sigma_t \nabla D_{face}(\hat z^{non\text{-}face}_{t}, a) * M_{face}.
\end{equation}

The video latent $\hat z_{t}^{*}$ is derived through classifier-based guidance, yielding the final predicted clean sample $\hat z_0$ via above iterative process.
\begin{equation}
\begin{aligned}
    \hat z_{t}^{*} & = \hat z_{t}^{face} + \hat z_{t}^{non\text{-}face} - \hat z_t,
    \label{eq:seq}
\end{aligned}
\end{equation}
\begin{equation}
    \hat z_{t-1}   = f_\theta(\hat z^{*}_{t}, a,t).
    \label{eq:back-infer}
\end{equation}

% \noindent (2) \textbf{Differential Guidance}
\subsubsection{\textbf{Differential Guidance (DG)}}~\label{sec:DG}
In our experiments, we observe that classifiers negatively affect the consistency of non-guided areas; for instance, the facial classifier reduces non-face alignment metrics (Table.~\ref{tab:component} ), and even mask conditions fails to completely isolate this effect. This is caused by Eq.\ref{eq:back-infer}, where gradients from face and non-face areas interact through the diffusion model.

% To address this issue, we propose differential guidance to manually compensate for the negative changes caused by classifiers in non-guided areas. Specifically, we modify Eq.\ref{eq:back-infer} in Sequential Guidance to:
To resolve this, we propose differential guidance to counteract negative changes caused by classifiers in non-guided areas. Specifically, we revise Eq.\ref{eq:back-infer} in Sequential Guidance to:
\begin{align}
  \hat z_{t-1} =  & f_\theta(\hat z_{t}^{*}, a, t) + \lambda_{diff}(f_\theta(\hat z_{t}^{non\text{-}face}, a, t) * M_{face} \notag\\ & + f_\theta(\hat z_{t}^{face}, a, t) * M_{non\text{-}face}). 
\end{align}
Here, $\lambda_{diff}$ is the weight that controls the differential strength. The formula can be understood as a compensation for the non-guided areas of the corresponding classifier. Differential Guidance, while slower than Sequential Guidance during inference, achieves superior generation quality (Table.~\ref{tab:main_result}).
The pseudocode for SG, DG, and the inference pipeline is detailed in Appendix B.4.2.
% \input{pcode/inference}

% \noindent \textbf{Long Video Generation. }

\subsection{CNAS Dataset}\label{sec:dataset}
% \zjn{To explain the necessity of constructing this dataset and the advantages it brings.} 
To address the scarcity of Chinese co-speech data, we construct the Chinese News Anchor Speech Dataset (CNAS). The dataset features single speakers primarily facing the camera, with above-waist perspectives and communication in Chinese. After preprocessing, 1,473 valid clips are obtained. More details are provided in Appendix C. This Chinese broadcasting dataset will be made publicly available for broader research.
% to provide high-quality data for broader research and exploration within the community.

%% file: secs/experiments.tex
\section{Experiments}

\input{tables/main_result}

\subsection{Experimental Settings}\label{sec:settings}
\noindent \textbf{Implement details. }
UniVDM is initialized from a pretrained video diffusion model~\cite{blattmann2023align}. During training with the PAR method, videos have a spatial resolution of 512×512 and a fixed length of 32 frames. We choose DDPM~\cite{ho2020denoising} as the noise scheduler with 1000 sampling steps. The confidence score threshold $\tau_j$ is set to 0.8, with a radius $r$ of 10. The hand/face weight hyperparameter $\omega_1$ is set to 10, while $\omega_2$ for the body region is set to 2.
In PCE classifiers training, negative samples are generated with a 30-step DDIM~\cite{song2020denoising} sampler. Facial boxes are extracted using MediaPipe~\cite{lugaresi2019mediapipe}. Both the non-face classifier and face classifier encoders are initialized with pre-trained $G$. For final inference, we use a 30-step DDIM sampler, applying classifier guidance only in the first 15 steps. 
% The guidance strengths for the face classifier, non-face classifier, and differential guidance ($\lambda_{face}$, $\lambda_{non\text{-}face}$, $\lambda_{diff}$) are set to 0.1, 1, and 0.25, respectively.
% We set $\lambda_{face}$ significantly lower than $\lambda_{non\text{-}ace}$ because experiments show that the face classifier tends to dominate the generation process more easily than the non-face classifier, and excessive $\lambda_{face}$ notably decreases video quality. The final video resolution is 256$\times$256. 
More implementation details can be found in Appendix B.
% (see Table~\ref{tab:face} in ablation study for details). 

\noindent \textbf{Datasets. }
The training data for our backbone network $G$ comes from the PATS dataset~\cite{ahuja2020style} and the CNAS dataset we propose. For fair comparison, we use the same training subset as S2G~\cite{he2024co}, which includes four talkers: Jon, Chemistry, Oliver, and Seth. Each speaker has 1,200 valid clips (without cutouts or camera movements), with clip lengths ranging from 4 to 15 seconds, at 25 fps, and training resolution uniformly adjusted to 512$\times$512. 90\% of the data is used for training and 10\% for evaluation. The CNAS dataset follows the same configuration as PATS.
% % ahuja2020no, ginosar2019learning, 
Positive audio-video samples for training the PCE classifiers are sourced from the processed PATS training set, while negative samples are generated by the backbone network $G$ pretrained for 60k steps, creating one-to-one paired positive and negative samples. Data augmentation strategies from Section~\ref{sec:pce} expand the training set, with audio and video lengths uniformly sampled across 30, 60, 90, and 120 frames. These augmentations yield 50k audio-video pairs, with a mask probability of 80\%.

% The positive audio-video samples for training the PCE classifiers come from our processed PATS training set described above, while negative samples are generated by the backbone network $G$ pretrained for 60k steps, forming one-to-one paired positive and negative samples. We expand the training set using the data augmentation strategies introduced in Section~\ref{sec:pce}, where the audio and video lengths are uniformly sampled among four frame lengths: 30, 60, 90, and 120. After these augmentation strategies, we obtain 50k pairs of audio-video samples. The mask probability is 80\%.

\noindent \textbf{Evaluation metrics. }1) Fréchet Gesture Distance (FGD)~\cite{yoon2020speech}; 2) Diversity (Div.); 3) Beat Alignment Score (BAS)~\cite{li2021learn} 4) Synchronization-C (Sync-C)~\cite{xu2024hallo}; 5) Fréchet Video Distance (FVD)~\cite{unterthiner2018towards}. The meanings of each metric are detailed in Appendix B.5.
% % the pose estimator 
% To evaluate the quality, diversity, and alignment between gestures and speech, we employ: 

\noindent \textbf{Baselines.} We compare our method with four baselines: 1) S2G~\cite{he2024co}, the latest SOTA in gesture video generation; 2) ANGIE~\cite{liu2022audio}; 3) SDT~\cite{qian2021speech}; and 4) MM-Diffusion~\cite{ruan2023mm}. All baselines are initialized with official weights and fine-tuned on the PATS and CNAS datasets.
% Except for ANGIE, which lacks open-source training code, the other 
% \begin{figure}[!tbp]
%     \centering
%     \includegraphics[width=.95\linewidth]{figures/show_1.png}
%     \caption{\label{fig:show} A qualitative comparison highlights flaws in earlier methods, marked with red circles and boxes, such as background inconsistencies, hand blurriness, and finger distortion. In contrast, our method produces high-quality videos free from these artifacts.}
% \end{figure}
\subsection{Quantitative Results}
The quantitative results reported in Table~\ref{tab:main_result} show that our method achieves the best performance in quality, alignment, and motion metrics. This demonstrates that our end-to-end Unified Diffusion Model can generate realistic motion videos with high speech consistency under the simultaneous influence of the PAR and PCE methods. Our approach using Differential Guidance outperforms other metrics except Div. compared to Sequential Guidance. Furthermore, even though our method primarily focuses on optimizing specific areas, the improvement in FVD strongly demonstrates the method's gain in overall quality.
%\zjn{Tables and figures should be placed near the corresponding text sections.}

\subsection{User Study}
Audio-driven human animation relies heavily on subjective perception over objective metrics. To evaluate our method's visual performance, we conducted a user study comparing videos generated by S2G, MM-Diffusion, SDT, and PAHA-DG. Ten videos were randomly selected from the PATS and CNAS test sets, and 16 participants evaluated them based on realism, diversity, speech-motion synchronization, and overall quality. Participants were instructed to ignore texture and facial expressions during motion evaluations. Preferences for each criterion were measured independently. As shown in Table~\ref{tab:user-study}, our method outperformed others across all criteria, especially in overall quality and synchronization, proving its ability to generate high-quality co-speech gesture videos while balancing motion and visual effects.

% \vspace{-10pt}
\input{tables/component}
\subsection{Ablation Study}
% We conduct an ablation study on the PATS. 
% test set, mainly focusing on the core components of the framework and the guidance hyperparameters.

\noindent \textbf{Core Components. }We identify three key components in the method: PAR, non-face classifier guidance, and face classifier guidance. Sequential Guidance is used for video generation during inference. Table~\ref{tab:component} shows that removing PAR worsens video quality (FVD +92.77) and reduces Diversity (-11.051). Face classifier guidance significantly affects lip alignment (Sync-C -14.9\%), while the non-face classifier primarily impacts gesture movement (FGD +19.5\%, BAS -16.6\%), aligning with expectations. Both classifiers improve alignment and motion metrics but slightly degrade FVD, with the face classifier having a stronger effect. 
% To mitigate this, Sequential Guidance applies the non-face classifier first and reduces the face classifier's guidance weight.
% \zjn{It is better to add qualitative visualizations for ablating each component.}

\noindent \textbf{Guidance Parameters. }We determine the guidance strength during inference via ablation experiments. The complete results for the non-face guidance weight $\lambda_{non\text{-}face}$, face guidance weight $\lambda_{face}$ and differential guidance weight $\lambda_{diff}$ are available in Appendix D. The final weight combination is set to $(\lambda_{non\text{-}face}=1, \lambda_{face}=0.1, \lambda_{diff}=0.25)$, achieving optimal performance.

\noindent \textbf{Guidance Time Rate. } 
Table~\ref{tab:time} compares the performance of PAHA-SG (\textit{Sequential Guidance}) and PAHA-DG (\textit{Differential Guidance}) at different proportions of guidance steps. A new metric, \textit{Time Cost (TC)}, measures the average time (in seconds) to generate a video segment (256x256 resolution, 30 steps), rounded to the nearest integer. In the paired values, the first represents PAHA-SG, and the second is PAHA-DG. The data shows that increasing classifier-guided steps raises time costs. While FVD worsens, other metrics improve. The best performance balance occurs as the guided step proportion increases from 0\% to 50\%, but further increases reduce performance.

\noindent \textbf{Inference Time. } As shown in Fig.~\ref{fig:teaser}, our method significantly outperforms the multi-stage co-speech gesture method in inference efficiency using the TC metric. While its inference time is comparable to ANGIE, our method delivers noticeably better results. Operating in the latent space further enhances its efficiency.

\noindent \textbf{Data Augmentation. }We also perform an ablation analysis of the data augmentation strategy, which demonstrates its effectiveness in improving all metrics (Table.~\ref{tab:component}).

\input{tables/user_study}

%% file: tables/main_result.tex
\definecolor{mygray2}{gray}{.8}
\renewcommand{\arraystretch}{1} % 设置行间距为1.5倍
\begin{table*}[htbp]
\centering
\caption{\label{tab:main_result}Quantitative results on the pats and cnas datasets. the best metrics are highlighted in \textbf{BLOD} and \underline{UNDERLINE} indicates the second-best. SG stands for \textbf{Sequential Guidance}, and DG stands for \textbf{Differential Guidance}.
}
\begin{tabular}{lccccc|ccccc}
\toprule
\multirow{2}{*}{Method} & \multicolumn{5}{c}{\raisebox{0.5ex}{PATS}} & \multicolumn{5}{c}{\raisebox{0.5ex}{CNAS}}   \\ \cline{2-11}
  & FVD $\downarrow$                  & BAS $\uparrow$                   & FGD $\downarrow$ & Sync-C $\uparrow$  & \multicolumn{1}{c|}{Div. $\uparrow$} & FVD $\downarrow$ & BAS $\uparrow$ & FGD $\downarrow$ & Sync-C $\uparrow$ & Div. $\uparrow$    \\ \midrule
% \rowcolor{mygray2} Ground Truth (GT) &   1883.64    &   0.1823    &  9.157   & 7.794  &   109.02 &   1326.75     &    0.1601    &  14.550   &  5.829   &    46.36    \\ \midrule 
SDT~\cite{qian2021speech}          & 2281.65                           & 0.1067                           & 22.966           & 5.337                           & 83.917    & 1592.48     & 0.0819   & 28.559   & 3.374     & 29.73         \\
ANGIE~\cite{liu2022audio}        & 2477.23                           & 0.0892                           & 42.081           & 4.927                           & 79.102    & /     &/   & /   & /     & /               \\
MM-Diffusion~\cite{ruan2023mm} & 2204.09                           & 0.1085                           & 23.782           & 5.348                           & 82.545      & 1531.21     & 0.0883   & 27.664   & 3.288     & 26.11                     \\
S2G~\cite{he2024co}          & 2118.41                           & 0.1227                           & 19.816           & 5.294                           & 85.623    & 1465.75     & 0.1087   & 27.523   & 3.452     & 32.65           \\
PAHA-SG (Ours) & \underline{2052.95}                           & \underline{0.1561}                           & \underline{15.021}           & \underline{6.183}                           & \textbf{100.917}   & \underline{1379.57}     & \underline{0.1241}   & \underline{24.439}   & \underline{3.81}     & \textbf{39.18}    \\
PAHA-DG (Ours) & \textbf{2048.75} & \textbf{0.1602} & \textbf{14.661} & \textbf{6.328} & \underline{98.235}    & \textbf{1362.44}   &\textbf{0.1275}   & \textbf{23.715}   & \textbf{4.092}     & \underline{37.54}                        \\ \bottomrule
\end{tabular}
\end{table*}

% \begin{table*}[tbp] 
\begin{table*}[tbp]  
\centering  
\caption{\label{tab:time}Ablation study regarding guidance time. in each number pair, the former is PAHA-SG and the latter is PAHA-DG. $^\star$ indicates the best of PAHA-SG, $^\bullet$ indicates the best of PAHA-DG.}    
\begin{tabular}{cccccccc}
\hline
Guidance Time Rate & FVD $\downarrow$     & BAS $\uparrow$       & FGD $\downarrow$     & Sync-C $\uparrow$    & Div. $\uparrow$  & TC $\downarrow$      \\ \hline
0\%             &      2047.53$^\star$\;/\;2047.53$^\bullet$               &         
 0.1346\;/\;0.1346           &         17.675\;/\;17.675               &     5.709\;/\;5.709                 &     90.804\;/\;90.804      &  85$^\star$\;/\;85$^\bullet$  \\
25\%             &       2050.66\;/\;2048.09          &      0.1423\;/\;0.1507        &       16.790\;/\;16.238            &      6.015\;/\;6.094               &    93.713\;/\;95.482  &     91\;/\;96    \\
50\%             &             2052.95\;/\;2048.75 & 0.1561\;/\;0.1602 & 15.021$^\star$\;/\;14.661$^\bullet$ & 6.183\;/\;6.328  & 99.604$^\star$\;/\;98.235$^\bullet$  &   98\;/\;112  \\
75\%      &       2056.03\;/\;2051.78        &    0.1662\;/\;0.1689       &      17.882\;/\;17.104         &                     6.306\;/\;6.451      &     96.122\;/\;97.318      &   107\;/\;129 \\
100\%    &   2062.95\;/\;2055.43        &    0.1674$^\star$\;/\;0.1711$^\bullet$         &  18.723\;/\;17.868      &         6.382$^\star$\;/\;6.546$^\bullet$       &           92.071\;/\;93.944        &    114\;/\;141      
\\ \hline
\end{tabular}
\end{table*}

%% file: tables/component.tex
% \begin{table}[htbp]
% \centering
% \caption{\label{tab:component}ABLATION STUDY RESULTS REGARDING CORE MODULES. BOLD INDICATES THE BEST AND UNDERLINE INDICATES THE SECOND. ``W/O' IS SHORT FOR ``WITHOUT'.}
% \begin{tabular}{lccccc}
% \toprule
% Name                         & FVD $\downarrow$ & BAS $\uparrow$  & FGD $\downarrow$ & Sync-C $\uparrow$ & Div. $\uparrow$  \\ \midrule
% w/o PAR                      & 2145.72          & 0.1497          & 15.452           & 5.977             & 89.866           \\
% w/o hand classifier & 2087.49          & 0.1302          & 17.957           & \textbf{6.583}    & 97.483           \\
% w/o face classifier & 2060.11          & 0.1554          & 15.181           & 5.726             & 99.154           \\ \midrule
% w/o augmentation & 2068.74          & 0.1548          & 15.279           & 6.126             & 99.604           \\ \midrule
% Ours                         & \textbf{2052.95} & \textbf{0.1561} & \textbf{15.021}  & 6.183             & \textbf{100.917} \\ \bottomrule
% \end{tabular}
% \end{table} 

\begin{table}[tp]  
\centering  
\caption{\label{tab:component}Ablation study regarding core modules. \textbf{BOLD} indicates the best ``w/o'' is short for ``without''.}  
\resizebox{\linewidth}{!}{
\begin{tabular}{lcccccc}  
\toprule  
Name                         & FVD $\downarrow$ & BAS $\uparrow$  & FGD $\downarrow$ & Sync-C $\uparrow$ & Div. $\uparrow$  \\ \midrule  
w/o PAR                      & 2145.72          & 0.1497          & 15.452           & 5.977             & 89.866           \\
w/o non-face classifier          & 2087.49          & 0.1302          & 17.957           & \textbf{6.583}    & 97.483           \\
w/o face classifier          & 2060.11          & 0.1554          & 15.181           & 5.726             & 99.154           \\ 
w/o PAR+two classifiers          & 2126.72          & 0.1197          & 20.052           & 5.227             & 83.772           \\ \midrule  
w/o augmentation             & 2068.74          & 0.1548          & 15.279           & 6.126             & 99.604           \\ \midrule  
Ours                         & \textbf{2052.95} & \textbf{0.1561} & \textbf{15.021}  & 6.183             & \textbf{100.917} \\ \bottomrule  
\end{tabular}  
}
\end{table}

%% file: tables/user_study.tex
\definecolor{mygray2}{gray}{.8}
\begin{table}[!tbp]
\centering
\caption{\label{tab:user-study} The preferred percentage of our method and the baselines in user study on the PATS and CNAS dataset.
}
\resizebox{\linewidth}{!}{ 
\begin{tabular}{lcccc}
\toprule
 Method   & {Realness $\uparrow$} & {Diversity $\uparrow$} & {Synchrony $\uparrow$} & {Overall quality $\uparrow$} \\
\midrule
% \rowcolor{mygray2} GT          &               94.00\%                          &      94.00\%       &               94.00\%              &                                93.00\%                \\ 
% \midrule
SDT~\cite{qian2021speech}          &       15.00\%                               &    8.20\%         &      10.00\%                       &         6.20\%                                       \\
MM-Diff~\cite{ruan2023mm}   &       8.60\%                                 &     13.60\%       &       11.80\%                       &         10.60\%                                       \\
S2G~\cite{he2024co}          &    12.60\%                                     &   15.60\%         &    13.60\%                          &    13.80\%                                            \\
PAHA-DG &         \textbf{63.80\%}               &    \textbf{62.60\%}            &   \textbf{64.60\%}                       &      \textbf{69.40\%}                                          \\ 
\bottomrule
\end{tabular}
}
\end{table}

%%{\linewidth}

%% file: secs/conclusion.tex
\section{Conclusion}
% we introduce PAHA, an end-to-end video generation framework based on diffusion models that creates high-quality audio-driven half-body human animations without requiring intermediate representation. We design a unified video diffusion model to process reference images and noisy videos simultaneously. During training, we employ Parts-Aware Re-weighting (PAR) for dynamic loss re-weighting in any region, allowing the model to focus more effectively on foreground generation and enhance visual quality. The Parts Consistency Enhancement (PCE) method employs self-distillation to sample video-audio pairs, training diffusion-based regional classifiers that produce alignment gradients during inference to direct the model's attention to motion-audio feature consistency in specific areas. Extensive experiments show that our framework generates visually appealing animations while ensuring temporal consistency between motion and audio, significantly surpassing existing methods. Additionally, we build the first Chinese broadcasting dataset CNAS in the field of co-speech gestures. We will open it up in the future to further promote the development of this field.
This paper introduces PAHA, an end-to-end diffusion-based framework for generating high-quality audio-driven half-body human animations without intermediate representations. The unified video diffusion model processes reference images and noisy videos simultaneously. During training, Parts-Aware Re-weighting (PAR) dynamically adjusts loss to focus on foreground regions, enhancing visual quality. The Parts Consistency Enhancement (PCE) method uses self-distillation to train diffusion-based regional classifiers, which provide alignment gradients during inference to ensure motion-audio consistency in specific areas. Experiments show that PAHA produces visually appealing, temporally consistent animations, surpassing existing methods. Additionally, we present CNAS, the first Chinese broadcasting dataset for co-speech gestures.

%% file: pcode/inference.tex
\begin{algorithm}
	%\textsl{}\setstretch{1.8}
	\renewcommand{\algorithmicrequire}{\textbf{Input:}}
	\renewcommand{\algorithmicensure}{\textbf{Output:}}
        \caption{Classifier-based Inference}
        \label{alg:inference}
	\begin{algorithmic}[1]
		\REQUIRE A reference image $\mathbf{I_{ref}}$, a driven-audio $\mathbf{a}$, the unified diffusion model $G_{\mathrm{PAR}}$ trained by PAR, the face classifier $D_{face}$, the non-face classifier $D_{non\text{-}face}$, timestep $t_{N-1}>t_{N-2}>...>t_1$, ODE solver $f_\theta$, noise scheduler $\alpha(t),\sigma(t)$, face mask $M_{face}$, non-face mask $M_{non\text{-}face}$, guidance weight $\lambda_{face},\lambda_{non\text{-}face},\lambda_{diff}$, guidance rate $\mathbf{r}$, decoder $\mathbf{D}$
        \ENSURE the generated co-speech video $\mathbf{v}$
        \STATE Sample Guassian Noise $z_T \sim \mathcal{N}(0, I)$
        \FOR {$n=N-1$ to $(N-1)(1-r)$}
           \STATE $\hat z_{t_n} = f_\theta(\hat z_{t_{n+1}}, a, t_{n+1})$
           \STATE $\begin{aligned}
            \hat z_{t_n}^{non\text{-}face} = & \hat z_{t_n} - \lambda_{non\text{-}face} \sigma_{t_n} \nabla D_{non\text{-}face}(\hat z_{t_n}, a) * \\  
             & M_{non\text{-}face}
            \end{aligned}$
            \STATE $\hat z_{t_n}^{face} = \hat z_{t_n} - \lambda_{face} \sigma_{t_n} \nabla D_{face}(\hat z^{non\text{-}face}_{t_n}, a) * M_{face}$
            \STATE $\hat z_{t_n}^{*} = \hat z_{t_n}^{face} + \hat z_{t_n}^{non\text{-}face} - \hat z_{t_n}$
            \IF{guidance name is \textit{\textbf{``Sequential Guidance (SG)''}}}
            \STATE $\hat z_{t_{n-1}}  = f_\theta(\hat z^{*}_{t_n}, a,t_n)$
            \ELSIF{guidance name is \textit{\textbf{``Differential Guidance (DG)''}}}
            \STATE $\begin{aligned}
            \hat z_{t_{n-1}}  = & f_\theta(\hat z_{t_n}^{*}, a, t_n) + \lambda_{diff}(f_\theta(\hat z_{t_n}^{non\text{-}face}, a, t_n) * M_{face} + \\
            & f_\theta(\hat z_{t_n}^{face}, a, t_n) * M_{non\text{-}face})
            \end{aligned}$
            \ENDIF 
        \ENDFOR
        % \State \;\;\;\;  $\hat z_{t_n} = f_\theta(\hat z^{*}_{t_{n+1}}, a, t_{n+1})$
	\end{algorithmic}  
\end{algorithm}

%% file: tables/hand_weight.tex
\begin{table}[tp]  
\centering  
\caption{\label{tab:hand}Ablation study regarding non-face guidance strength. \textbf{BOLD} indicates the best.}  
\begin{tabular}{lccccc}
\hline
$\lambda_{non\text{-}face}$ & FVD $\downarrow$     & BAS $\uparrow$       & FGD $\downarrow$     & Sync-C $\uparrow$    & Div. $\uparrow$      \\ \hline
0.8             &       2049.91        &    0.1578       &      \textbf{14.107}       &       \textbf{6.436}      &     95.271         \\
0.9             &       2047.68        &    0.1591        &      14.393           &      6.390        &       96.648             \\
1.0            &       2045.75    &       \textbf{0.1602}      & 14.661    &   6.328  & 98.235        \\
1.1             &       2044.09      &     0.1599      &     14.917      &        6.247        &      \textbf{101.443}      \\
1.2             &     \textbf{2042.87}      &        0.1596        &   15.265      &      6.146       &    100.143        \\ \hline
\end{tabular}
\end{table}

%% file: tables/face_weight.tex
\begin{table}[tp]  
\centering  
\caption{\label{tab:face}Ablation study regarding face guidance strength. \textbf{BOLD} indicates the best.}  
\begin{tabular}{lccccc}
\hline
$\lambda_{face}$ & FVD $\downarrow$     & BAS $\uparrow$       & FGD $\downarrow$     & Sync-C $\uparrow$    & Div. $\uparrow$      \\ \hline
0.05             &       2041.67      &      \textbf{0.1629}             &         14.682         &        6.197        &   \textbf{100.267}    \\
0.10             &             \textbf{2045.75} & 0.1602 & \textbf{14.661} & \textbf{6.328}  & 98.235       \\
0.15             &         2047.85             &     0.1594               &       14.736             &       6.416               &    96.563    \\
0.20             &          2050.99            &        0.1581          &         14.799             &     6.480                 &      94.782   \\
0.25             &        2056.39         &           0.1571        &       14.882      &          6.529      &                93.551          \\ \hline
\end{tabular}
\end{table}

%% file: tables/diff_weight.tex
\begin{table}[tp]  
\centering  
\caption{\label{tab:diff}Ablation study regarding differential guidance strength. \textbf{BOLD} indicates the best.}  
\begin{tabular}{lccccc}
\hline
$\lambda_{diff}$ & FVD $\downarrow$     & BAS $\uparrow$       & FGD $\downarrow$     & Sync-C $\uparrow$    & Div. $\uparrow$      \\ \hline
0.15             &        2053.63          &   0.1584          &     14.805           &      6.257           &   \textbf{99.580}     \\
0.20             &        2048.11        &     0.1596          &     14.739           &      6.301           &   98.742    \\
0.25             &            2045.75 & \textbf{0.1602} & 14.661 & \textbf{6.328}  & 98.235        \\
0.30             &           \textbf{2043.98}        &    0.1593         &      14.582       &      6.296     &   97.611      \\
0.35            &            2045.53    &    0.1582   &   \textbf{14.355}  &  6.235   &   97.078      \\ \hline
\end{tabular}
\end{table}

%% file: arxiv-main.bbl

\begin{thebibliography}{61}

%%% the Web address.

\ifx \showCODEN    \undefined \def \showCODEN     #1{\unskip}     \fi
\ifx \showISBNx    \undefined \def \showISBNx     #1{\unskip}     \fi
\ifx \showISBNxiii \undefined \def \showISBNxiii  #1{\unskip}     \fi
\ifx \showISSN     \undefined \def \showISSN      #1{\unskip}     \fi
\ifx \showLCCN     \undefined \def \showLCCN      #1{\unskip}     \fi
\ifx \shownote     \undefined \def \shownote      #1{#1}          \fi
\ifx \showarticletitle \undefined \def \showarticletitle #1{#1}   \fi
\ifx \showURL      \undefined \def \showURL       {\relax}        \fi
% invisible to TeX
\providecommand\bibfield[2]{#2}
\providecommand\bibinfo[2]{#2}
\providecommand\natexlab[1]{#1}
\providecommand\showeprint[2][]{arXiv:#2}

\bibitem[Ahuja et~al\mbox{.}(2020)]%
        {ahuja2020style}
\bibfield{author}{\bibinfo{person}{Chaitanya Ahuja}, \bibinfo{person}{Dong~Won Lee}, \bibinfo{person}{Yukiko~I Nakano}, {and} \bibinfo{person}{Louis-Philippe Morency}.} \bibinfo{year}{2020}\natexlab{}.
\newblock \showarticletitle{Style transfer for co-speech gesture animation: A multi-speaker conditional-mixture approach}. In \bibinfo{booktitle}{\emph{Computer Vision--ECCV 2020: 16th European Conference, Glasgow, UK, August 23--28, 2020, Proceedings, Part XVIII 16}}. Springer, \bibinfo{pages}{248--265}.
\newblock


\bibitem[Blattmann et~al\mbox{.}(2023)]%
        {blattmann2023align}
\bibfield{author}{\bibinfo{person}{Andreas Blattmann}, \bibinfo{person}{Robin Rombach}, \bibinfo{person}{Huan Ling}, \bibinfo{person}{Tim Dockhorn}, \bibinfo{person}{Seung~Wook Kim}, \bibinfo{person}{Sanja Fidler}, {and} \bibinfo{person}{Karsten Kreis}.} \bibinfo{year}{2023}\natexlab{}.
\newblock \showarticletitle{Align your latents: High-resolution video synthesis with latent diffusion models}. In \bibinfo{booktitle}{\emph{Proceedings of the IEEE/CVF Conference on Computer Vision and Pattern Recognition}}. \bibinfo{pages}{22563--22575}.
\newblock


\bibitem[Cai et~al\mbox{.}(2024)]%
        {cai2024diffusion}
\bibfield{author}{\bibinfo{person}{Shengqu Cai}, \bibinfo{person}{Eric Chan}, \bibinfo{person}{Yunzhi Zhang}, \bibinfo{person}{Leonidas Guibas}, \bibinfo{person}{Jiajun Wu}, {and} \bibinfo{person}{Gordon Wetzstein}.} \bibinfo{year}{2024}\natexlab{}.
\newblock \showarticletitle{Diffusion Self-Distillation for Zero-Shot Customized Image Generation}.
\newblock \bibinfo{journal}{\emph{arXiv preprint arXiv:2411.18616}} (\bibinfo{year}{2024}).
\newblock


\bibitem[Chatziagapi et~al\mbox{.}(2024)]%
        {chatziagapi2024talkinnerf}
\bibfield{author}{\bibinfo{person}{Aggelina Chatziagapi}, \bibinfo{person}{Bindita Chaudhuri}, \bibinfo{person}{Amit Kumar}, \bibinfo{person}{Rakesh Ranjan}, \bibinfo{person}{Dimitris Samaras}, {and} \bibinfo{person}{Nikolaos Sarafianos}.} \bibinfo{year}{2024}\natexlab{}.
\newblock \showarticletitle{TalkinNeRF: Animatable Neural Fields for Full-Body Talking Humans}.
\newblock \bibinfo{journal}{\emph{arXiv preprint arXiv:2409.16666}} (\bibinfo{year}{2024}).
\newblock


\bibitem[Chen et~al\mbox{.}(2024)]%
        {chen2024videocrafter2}
\bibfield{author}{\bibinfo{person}{Haoxin Chen}, \bibinfo{person}{Yong Zhang}, \bibinfo{person}{Xiaodong Cun}, \bibinfo{person}{Menghan Xia}, \bibinfo{person}{Xintao Wang}, \bibinfo{person}{Chao Weng}, {and} \bibinfo{person}{Ying Shan}.} \bibinfo{year}{2024}\natexlab{}.
\newblock \showarticletitle{Videocrafter2: Overcoming data limitations for high-quality video diffusion models}. In \bibinfo{booktitle}{\emph{Proceedings of the IEEE/CVF Conference on Computer Vision and Pattern Recognition}}. \bibinfo{pages}{7310--7320}.
\newblock


\bibitem[Corona et~al\mbox{.}(2024)]%
        {corona2024vlogger}
\bibfield{author}{\bibinfo{person}{Enric Corona}, \bibinfo{person}{Andrei Zanfir}, \bibinfo{person}{Eduard~Gabriel Bazavan}, \bibinfo{person}{Nikos Kolotouros}, \bibinfo{person}{Thiemo Alldieck}, {and} \bibinfo{person}{Cristian Sminchisescu}.} \bibinfo{year}{2024}\natexlab{}.
\newblock \showarticletitle{VLOGGER: Multimodal diffusion for embodied avatar synthesis}.
\newblock \bibinfo{journal}{\emph{arXiv preprint arXiv:2403.08764}} (\bibinfo{year}{2024}).
\newblock


\bibitem[Ginosar et~al\mbox{.}(2019)]%
        {ginosar2019learning}
\bibfield{author}{\bibinfo{person}{Shiry Ginosar}, \bibinfo{person}{Amir Bar}, \bibinfo{person}{Gefen Kohavi}, \bibinfo{person}{Caroline Chan}, \bibinfo{person}{Andrew Owens}, {and} \bibinfo{person}{Jitendra Malik}.} \bibinfo{year}{2019}\natexlab{}.
\newblock \showarticletitle{Learning individual styles of conversational gesture}. In \bibinfo{booktitle}{\emph{Proceedings of the IEEE/CVF Conference on Computer Vision and Pattern Recognition}}. \bibinfo{pages}{3497--3506}.
\newblock


\bibitem[Guo et~al\mbox{.}(2024)]%
        {guo2024liveportrait}
\bibfield{author}{\bibinfo{person}{Jianzhu Guo}, \bibinfo{person}{Dingyun Zhang}, \bibinfo{person}{Xiaoqiang Liu}, \bibinfo{person}{Zhizhou Zhong}, \bibinfo{person}{Yuan Zhang}, \bibinfo{person}{Pengfei Wan}, {and} \bibinfo{person}{Di Zhang}.} \bibinfo{year}{2024}\natexlab{}.
\newblock \showarticletitle{Liveportrait: Efficient portrait animation with stitching and retargeting control}.
\newblock \bibinfo{journal}{\emph{arXiv preprint arXiv:2407.03168}} (\bibinfo{year}{2024}).
\newblock


\bibitem[Guo et~al\mbox{.}(2023)]%
        {guo2023animatediff}
\bibfield{author}{\bibinfo{person}{Yuwei Guo}, \bibinfo{person}{Ceyuan Yang}, \bibinfo{person}{Anyi Rao}, \bibinfo{person}{Zhengyang Liang}, \bibinfo{person}{Yaohui Wang}, \bibinfo{person}{Yu Qiao}, \bibinfo{person}{Maneesh Agrawala}, \bibinfo{person}{Dahua Lin}, {and} \bibinfo{person}{Bo Dai}.} \bibinfo{year}{2023}\natexlab{}.
\newblock \showarticletitle{Animatediff: Animate your personalized text-to-image diffusion models without specific tuning}.
\newblock \bibinfo{journal}{\emph{arXiv preprint arXiv:2307.04725}} (\bibinfo{year}{2023}).
\newblock


\bibitem[He et~al\mbox{.}(2024)]%
        {he2024co}
\bibfield{author}{\bibinfo{person}{Xu He}, \bibinfo{person}{Qiaochu Huang}, \bibinfo{person}{Zhensong Zhang}, \bibinfo{person}{Zhiwei Lin}, \bibinfo{person}{Zhiyong Wu}, \bibinfo{person}{Sicheng Yang}, \bibinfo{person}{Minglei Li}, \bibinfo{person}{Zhiyi Chen}, \bibinfo{person}{Songcen Xu}, {and} \bibinfo{person}{Xiaofei Wu}.} \bibinfo{year}{2024}\natexlab{}.
\newblock \showarticletitle{Co-Speech Gesture Video Generation via Motion-Decoupled Diffusion Model}. In \bibinfo{booktitle}{\emph{Proceedings of the IEEE/CVF Conference on Computer Vision and Pattern Recognition}}. \bibinfo{pages}{2263--2273}.
\newblock


\bibitem[Ho et~al\mbox{.}(2020)]%
        {ho2020denoising}
\bibfield{author}{\bibinfo{person}{Jonathan Ho}, \bibinfo{person}{Ajay Jain}, {and} \bibinfo{person}{Pieter Abbeel}.} \bibinfo{year}{2020}\natexlab{}.
\newblock \showarticletitle{Denoising diffusion probabilistic models}.
\newblock \bibinfo{journal}{\emph{Advances in neural information processing systems}}  \bibinfo{volume}{33} (\bibinfo{year}{2020}), \bibinfo{pages}{6840--6851}.
\newblock


\bibitem[Ho et~al\mbox{.}(2022)]%
        {ho2022video}
\bibfield{author}{\bibinfo{person}{Jonathan Ho}, \bibinfo{person}{Tim Salimans}, \bibinfo{person}{Alexey Gritsenko}, \bibinfo{person}{William Chan}, \bibinfo{person}{Mohammad Norouzi}, {and} \bibinfo{person}{David~J Fleet}.} \bibinfo{year}{2022}\natexlab{}.
\newblock \showarticletitle{Video diffusion models}.
\newblock \bibinfo{journal}{\emph{Advances in Neural Information Processing Systems}}  \bibinfo{volume}{35} (\bibinfo{year}{2022}), \bibinfo{pages}{8633--8646}.
\newblock


\bibitem[Hogue et~al\mbox{.}(2024)]%
        {hogue2024diffted}
\bibfield{author}{\bibinfo{person}{Steven Hogue}, \bibinfo{person}{Chenxu Zhang}, \bibinfo{person}{Hamza Daruger}, \bibinfo{person}{Yapeng Tian}, {and} \bibinfo{person}{Xiaohu Guo}.} \bibinfo{year}{2024}\natexlab{}.
\newblock \showarticletitle{DiffTED: One-shot Audio-driven TED Talk Video Generation with Diffusion-based Co-speech Gestures}. In \bibinfo{booktitle}{\emph{Proceedings of the IEEE/CVF Conference on Computer Vision and Pattern Recognition}}. \bibinfo{pages}{1922--1931}.
\newblock


\bibitem[Hu(2024)]%
        {hu2024animate}
\bibfield{author}{\bibinfo{person}{Li Hu}.} \bibinfo{year}{2024}\natexlab{}.
\newblock \showarticletitle{Animate anyone: Consistent and controllable image-to-video synthesis for character animation}. In \bibinfo{booktitle}{\emph{Proceedings of the IEEE/CVF Conference on Computer Vision and Pattern Recognition}}. \bibinfo{pages}{8153--8163}.
\newblock


\bibitem[Huang et~al\mbox{.}(2024)]%
        {huang2024make}
\bibfield{author}{\bibinfo{person}{Ziyao Huang}, \bibinfo{person}{Fan Tang}, \bibinfo{person}{Yong Zhang}, \bibinfo{person}{Xiaodong Cun}, \bibinfo{person}{Juan Cao}, \bibinfo{person}{Jintao Li}, {and} \bibinfo{person}{Tong-Yee Lee}.} \bibinfo{year}{2024}\natexlab{}.
\newblock \showarticletitle{Make-Your-Anchor: A Diffusion-based 2D Avatar Generation Framework}. In \bibinfo{booktitle}{\emph{Proceedings of the IEEE/CVF Conference on Computer Vision and Pattern Recognition}}. \bibinfo{pages}{6997--7006}.
\newblock


\bibitem[Kay et~al\mbox{.}(2017)]%
        {kay2017kinetics}
\bibfield{author}{\bibinfo{person}{Will Kay}, \bibinfo{person}{Joao Carreira}, \bibinfo{person}{Karen Simonyan}, \bibinfo{person}{Brian Zhang}, \bibinfo{person}{Chloe Hillier}, \bibinfo{person}{Sudheendra Vijayanarasimhan}, \bibinfo{person}{Fabio Viola}, \bibinfo{person}{Tim Green}, \bibinfo{person}{Trevor Back}, \bibinfo{person}{Paul Natsev}, {et~al\mbox{.}}} \bibinfo{year}{2017}\natexlab{}.
\newblock \showarticletitle{The kinetics human action video dataset}.
\newblock \bibinfo{journal}{\emph{arXiv preprint arXiv:1705.06950}} (\bibinfo{year}{2017}).
\newblock


\bibitem[Li et~al\mbox{.}(2021)]%
        {li2021learn}
\bibfield{author}{\bibinfo{person}{Ruilong Li}, \bibinfo{person}{Shan Yang}, \bibinfo{person}{David~A Ross}, {and} \bibinfo{person}{Angjoo Kanazawa}.} \bibinfo{year}{2021}\natexlab{}.
\newblock \showarticletitle{Learn to Dance with AIST++: Music Conditioned 3D Dance Generation. CoRR abs/2101.08779 (2021)}.
\newblock \bibinfo{journal}{\emph{arXiv preprint arXiv:2101.08779}} (\bibinfo{year}{2021}).
\newblock


\bibitem[Lin et~al\mbox{.}(2024)]%
        {lin2024sdxl}
\bibfield{author}{\bibinfo{person}{Shanchuan Lin}, \bibinfo{person}{Anran Wang}, {and} \bibinfo{person}{Xiao Yang}.} \bibinfo{year}{2024}\natexlab{}.
\newblock \showarticletitle{Sdxl-lightning: Progressive adversarial diffusion distillation}.
\newblock \bibinfo{journal}{\emph{arXiv preprint arXiv:2402.13929}} (\bibinfo{year}{2024}).
\newblock


\bibitem[Liu et~al\mbox{.}(2024)]%
        {liu2024anitalker}
\bibfield{author}{\bibinfo{person}{Tao Liu}, \bibinfo{person}{Feilong Chen}, \bibinfo{person}{Shuai Fan}, \bibinfo{person}{Chenpeng Du}, \bibinfo{person}{Qi Chen}, \bibinfo{person}{Xie Chen}, {and} \bibinfo{person}{Kai Yu}.} \bibinfo{year}{2024}\natexlab{}.
\newblock \showarticletitle{Anitalker: animate vivid and diverse talking faces through identity-decoupled facial motion encoding}. In \bibinfo{booktitle}{\emph{Proceedings of the 32nd ACM International Conference on Multimedia}}. \bibinfo{pages}{6696--6705}.
\newblock


\bibitem[Liu et~al\mbox{.}(2022a)]%
        {liu2022audio}
\bibfield{author}{\bibinfo{person}{Xian Liu}, \bibinfo{person}{Qianyi Wu}, \bibinfo{person}{Hang Zhou}, \bibinfo{person}{Yuanqi Du}, \bibinfo{person}{Wayne Wu}, \bibinfo{person}{Dahua Lin}, {and} \bibinfo{person}{Ziwei Liu}.} \bibinfo{year}{2022}\natexlab{a}.
\newblock \showarticletitle{Audio-driven co-speech gesture video generation}.
\newblock \bibinfo{journal}{\emph{Advances in Neural Information Processing Systems}}  \bibinfo{volume}{35} (\bibinfo{year}{2022}), \bibinfo{pages}{21386--21399}.
\newblock


\bibitem[Liu et~al\mbox{.}(2022b)]%
        {liu2022learning}
\bibfield{author}{\bibinfo{person}{Xian Liu}, \bibinfo{person}{Qianyi Wu}, \bibinfo{person}{Hang Zhou}, \bibinfo{person}{Yinghao Xu}, \bibinfo{person}{Rui Qian}, \bibinfo{person}{Xinyi Lin}, \bibinfo{person}{Xiaowei Zhou}, \bibinfo{person}{Wayne Wu}, \bibinfo{person}{Bo Dai}, {and} \bibinfo{person}{Bolei Zhou}.} \bibinfo{year}{2022}\natexlab{b}.
\newblock \showarticletitle{Learning hierarchical cross-modal association for co-speech gesture generation}. In \bibinfo{booktitle}{\emph{Proceedings of the IEEE/CVF Conference on Computer Vision and Pattern Recognition}}. \bibinfo{pages}{10462--10472}.
\newblock


\bibitem[Loshchilov(2017)]%
        {loshchilov2017decoupled}
\bibfield{author}{\bibinfo{person}{I Loshchilov}.} \bibinfo{year}{2017}\natexlab{}.
\newblock \showarticletitle{Decoupled weight decay regularization}.
\newblock \bibinfo{journal}{\emph{arXiv preprint arXiv:1711.05101}} (\bibinfo{year}{2017}).
\newblock


\bibitem[Lugaresi et~al\mbox{.}(2019)]%
        {lugaresi2019mediapipe}
\bibfield{author}{\bibinfo{person}{Camillo Lugaresi}, \bibinfo{person}{Jiuqiang Tang}, \bibinfo{person}{Hadon Nash}, \bibinfo{person}{Chris McClanahan}, \bibinfo{person}{Esha Uboweja}, \bibinfo{person}{Michael Hays}, \bibinfo{person}{Fan Zhang}, \bibinfo{person}{Chuo-Ling Chang}, \bibinfo{person}{Ming~Guang Yong}, \bibinfo{person}{Juhyun Lee}, {et~al\mbox{.}}} \bibinfo{year}{2019}\natexlab{}.
\newblock \showarticletitle{Mediapipe: A framework for building perception pipelines}.
\newblock \bibinfo{journal}{\emph{arXiv preprint arXiv:1906.08172}} (\bibinfo{year}{2019}).
\newblock


\bibitem[Mahapatra et~al\mbox{.}(2025)]%
        {mahapatra2025co}
\bibfield{author}{\bibinfo{person}{Aniruddha Mahapatra}, \bibinfo{person}{Richa Mishra}, \bibinfo{person}{Renda Li}, \bibinfo{person}{Ziyi Chen}, \bibinfo{person}{Boyang Ding}, \bibinfo{person}{Shoulei Wang}, \bibinfo{person}{Jun-Yan Zhu}, \bibinfo{person}{Peng Chang}, \bibinfo{person}{Mei Han}, {and} \bibinfo{person}{Jing Xiao}.} \bibinfo{year}{2025}\natexlab{}.
\newblock \showarticletitle{Co-speech Gesture Video Generation with 3D Human Meshes}. In \bibinfo{booktitle}{\emph{European Conference on Computer Vision}}. Springer, \bibinfo{pages}{172--189}.
\newblock


\bibitem[Mughal et~al\mbox{.}(2024)]%
        {mughal2024convofusion}
\bibfield{author}{\bibinfo{person}{Muhammad~Hamza Mughal}, \bibinfo{person}{Rishabh Dabral}, \bibinfo{person}{Ikhsanul Habibie}, \bibinfo{person}{Lucia Donatelli}, \bibinfo{person}{Marc Habermann}, {and} \bibinfo{person}{Christian Theobalt}.} \bibinfo{year}{2024}\natexlab{}.
\newblock \showarticletitle{Convofusion: Multi-modal conversational diffusion for co-speech gesture synthesis}. In \bibinfo{booktitle}{\emph{Proceedings of the IEEE/CVF Conference on Computer Vision and Pattern Recognition}}. \bibinfo{pages}{1388--1398}.
\newblock


\bibitem[Nazarieh et~al\mbox{.}(2024)]%
        {nazarieh2024survey}
\bibfield{author}{\bibinfo{person}{Fatemeh Nazarieh}, \bibinfo{person}{Zhenhua Feng}, \bibinfo{person}{Muhammad Awais}, \bibinfo{person}{Wenwu Wang}, {and} \bibinfo{person}{Josef Kittler}.} \bibinfo{year}{2024}\natexlab{}.
\newblock \showarticletitle{A Survey of Cross-Modal Visual Content Generation}.
\newblock \bibinfo{journal}{\emph{IEEE Transactions on Circuits and Systems for Video Technology}} (\bibinfo{year}{2024}).
\newblock


\bibitem[Oquab et~al\mbox{.}(2023)]%
        {oquab2023dinov2}
\bibfield{author}{\bibinfo{person}{Maxime Oquab}, \bibinfo{person}{Timoth{\'e}e Darcet}, \bibinfo{person}{Th{\'e}o Moutakanni}, \bibinfo{person}{Huy Vo}, \bibinfo{person}{Marc Szafraniec}, \bibinfo{person}{Vasil Khalidov}, \bibinfo{person}{Pierre Fernandez}, \bibinfo{person}{Daniel Haziza}, \bibinfo{person}{Francisco Massa}, \bibinfo{person}{Alaaeldin El-Nouby}, {et~al\mbox{.}}} \bibinfo{year}{2023}\natexlab{}.
\newblock \showarticletitle{Dinov2: Learning robust visual features without supervision}.
\newblock \bibinfo{journal}{\emph{arXiv preprint arXiv:2304.07193}} (\bibinfo{year}{2023}).
\newblock


\bibitem[Peng et~al\mbox{.}(2024)]%
        {peng2024controlnext}
\bibfield{author}{\bibinfo{person}{Bohao Peng}, \bibinfo{person}{Jian Wang}, \bibinfo{person}{Yuechen Zhang}, \bibinfo{person}{Wenbo Li}, \bibinfo{person}{Ming-Chang Yang}, {and} \bibinfo{person}{Jiaya Jia}.} \bibinfo{year}{2024}\natexlab{}.
\newblock \showarticletitle{Controlnext: Powerful and efficient control for image and video generation}.
\newblock \bibinfo{journal}{\emph{arXiv preprint arXiv:2408.06070}} (\bibinfo{year}{2024}).
\newblock


\bibitem[Qian et~al\mbox{.}(2021)]%
        {qian2021speech}
\bibfield{author}{\bibinfo{person}{Shenhan Qian}, \bibinfo{person}{Zhi Tu}, \bibinfo{person}{Yihao Zhi}, \bibinfo{person}{Wen Liu}, {and} \bibinfo{person}{Shenghua Gao}.} \bibinfo{year}{2021}\natexlab{}.
\newblock \showarticletitle{Speech drives templates: Co-speech gesture synthesis with learned templates}. In \bibinfo{booktitle}{\emph{Proceedings of the IEEE/CVF International Conference on Computer Vision}}. \bibinfo{pages}{11077--11086}.
\newblock


\bibitem[Ronneberger et~al\mbox{.}(2015)]%
        {ronneberger2015u}
\bibfield{author}{\bibinfo{person}{Olaf Ronneberger}, \bibinfo{person}{Philipp Fischer}, {and} \bibinfo{person}{Thomas Brox}.} \bibinfo{year}{2015}\natexlab{}.
\newblock \showarticletitle{U-net: Convolutional networks for biomedical image segmentation}. In \bibinfo{booktitle}{\emph{MICCAI}}.
\newblock


\bibitem[Ruan et~al\mbox{.}(2023)]%
        {ruan2023mm}
\bibfield{author}{\bibinfo{person}{Ludan Ruan}, \bibinfo{person}{Yiyang Ma}, \bibinfo{person}{Huan Yang}, \bibinfo{person}{Huiguo He}, \bibinfo{person}{Bei Liu}, \bibinfo{person}{Jianlong Fu}, \bibinfo{person}{Nicholas~Jing Yuan}, \bibinfo{person}{Qin Jin}, {and} \bibinfo{person}{Baining Guo}.} \bibinfo{year}{2023}\natexlab{}.
\newblock \showarticletitle{MM-Diffusion: Learning multi-modal diffusion models for joint audio and video generation}. In \bibinfo{booktitle}{\emph{Proceedings of the IEEE/CVF Conference on Computer Vision and Pattern Recognition}}. \bibinfo{pages}{10219--10228}.
\newblock


\bibitem[Sauer et~al\mbox{.}(2025)]%
        {sauer2025adversarial}
\bibfield{author}{\bibinfo{person}{Axel Sauer}, \bibinfo{person}{Dominik Lorenz}, \bibinfo{person}{Andreas Blattmann}, {and} \bibinfo{person}{Robin Rombach}.} \bibinfo{year}{2025}\natexlab{}.
\newblock \showarticletitle{Adversarial diffusion distillation}. In \bibinfo{booktitle}{\emph{European Conference on Computer Vision}}. Springer, \bibinfo{pages}{87--103}.
\newblock


\bibitem[Schneider et~al\mbox{.}(2019)]%
        {schneider2019wav2vec}
\bibfield{author}{\bibinfo{person}{Steffen Schneider}, \bibinfo{person}{Alexei Baevski}, \bibinfo{person}{Ronan Collobert}, {and} \bibinfo{person}{Michael Auli}.} \bibinfo{year}{2019}\natexlab{}.
\newblock \showarticletitle{wav2vec: Unsupervised pre-training for speech recognition}.
\newblock \bibinfo{journal}{\emph{arXiv preprint arXiv:1904.05862}} (\bibinfo{year}{2019}).
\newblock


\bibitem[Singer et~al\mbox{.}(2022)]%
        {singer2022make}
\bibfield{author}{\bibinfo{person}{Uriel Singer}, \bibinfo{person}{Adam Polyak}, \bibinfo{person}{Thomas Hayes}, \bibinfo{person}{Xi Yin}, \bibinfo{person}{Jie An}, \bibinfo{person}{Songyang Zhang}, \bibinfo{person}{Qiyuan Hu}, \bibinfo{person}{Harry Yang}, \bibinfo{person}{Oron Ashual}, \bibinfo{person}{Oran Gafni}, {et~al\mbox{.}}} \bibinfo{year}{2022}\natexlab{}.
\newblock \showarticletitle{Make-a-video: Text-to-video generation without text-video data}.
\newblock \bibinfo{journal}{\emph{arXiv preprint arXiv:2209.14792}} (\bibinfo{year}{2022}).
\newblock


\bibitem[Sohl-Dickstein et~al\mbox{.}(2015)]%
        {sohl2015deep}
\bibfield{author}{\bibinfo{person}{Jascha Sohl-Dickstein}, \bibinfo{person}{Eric Weiss}, \bibinfo{person}{Niru Maheswaranathan}, {and} \bibinfo{person}{Surya Ganguli}.} \bibinfo{year}{2015}\natexlab{}.
\newblock \showarticletitle{Deep unsupervised learning using nonequilibrium thermodynamics}. In \bibinfo{booktitle}{\emph{International conference on machine learning}}. PMLR, \bibinfo{pages}{2256--2265}.
\newblock


\bibitem[Song et~al\mbox{.}(2020)]%
        {song2020denoising}
\bibfield{author}{\bibinfo{person}{Jiaming Song}, \bibinfo{person}{Chenlin Meng}, {and} \bibinfo{person}{Stefano Ermon}.} \bibinfo{year}{2020}\natexlab{}.
\newblock \showarticletitle{Denoising diffusion implicit models}.
\newblock \bibinfo{journal}{\emph{arXiv preprint arXiv:2010.02502}} (\bibinfo{year}{2020}).
\newblock


\bibitem[Song et~al\mbox{.}(2022)]%
        {song2022audio}
\bibfield{author}{\bibinfo{person}{Linsen Song}, \bibinfo{person}{Wayne Wu}, \bibinfo{person}{Chaoyou Fu}, \bibinfo{person}{Chen~Change Loy}, {and} \bibinfo{person}{Ran He}.} \bibinfo{year}{2022}\natexlab{}.
\newblock \showarticletitle{Audio-driven dubbing for user generated contents via style-aware semi-parametric synthesis}.
\newblock \bibinfo{journal}{\emph{IEEE Transactions on Circuits and Systems for Video Technology}} \bibinfo{volume}{33}, \bibinfo{number}{3} (\bibinfo{year}{2022}), \bibinfo{pages}{1247--1261}.
\newblock


\bibitem[Su et~al\mbox{.}(2024)]%
        {su2024roformer}
\bibfield{author}{\bibinfo{person}{Jianlin Su}, \bibinfo{person}{Murtadha Ahmed}, \bibinfo{person}{Yu Lu}, \bibinfo{person}{Shengfeng Pan}, \bibinfo{person}{Wen Bo}, {and} \bibinfo{person}{Yunfeng Liu}.} \bibinfo{year}{2024}\natexlab{}.
\newblock \showarticletitle{Roformer: Enhanced transformer with rotary position embedding}.
\newblock \bibinfo{journal}{\emph{Neurocomputing}}  \bibinfo{volume}{568} (\bibinfo{year}{2024}), \bibinfo{pages}{127063}.
\newblock


\bibitem[Taud and Mas(2018)]%
        {taud2018multilayer}
\bibfield{author}{\bibinfo{person}{Hind Taud} {and} \bibinfo{person}{Jean-Franccois Mas}.} \bibinfo{year}{2018}\natexlab{}.
\newblock \showarticletitle{Multilayer perceptron (MLP)}.
\newblock \bibinfo{journal}{\emph{Geomatic approaches for modeling land change scenarios}} (\bibinfo{year}{2018}), \bibinfo{pages}{451--455}.
\newblock


\bibitem[Tian et~al\mbox{.}(2025)]%
        {tian2025emo}
\bibfield{author}{\bibinfo{person}{Linrui Tian}, \bibinfo{person}{Qi Wang}, \bibinfo{person}{Bang Zhang}, {and} \bibinfo{person}{Liefeng Bo}.} \bibinfo{year}{2025}\natexlab{}.
\newblock \showarticletitle{EMO: Emote Portrait Alive Generating Expressive Portrait Videos with Audio2Video Diffusion Model Under Weak Conditions}. In \bibinfo{booktitle}{\emph{European Conference on Computer Vision}}. Springer, \bibinfo{pages}{244--260}.
\newblock


\bibitem[Unterthiner et~al\mbox{.}(2018)]%
        {unterthiner2018towards}
\bibfield{author}{\bibinfo{person}{Thomas Unterthiner}, \bibinfo{person}{Sjoerd Van~Steenkiste}, \bibinfo{person}{Karol Kurach}, \bibinfo{person}{Raphael Marinier}, \bibinfo{person}{Marcin Michalski}, {and} \bibinfo{person}{Sylvain Gelly}.} \bibinfo{year}{2018}\natexlab{}.
\newblock \showarticletitle{Towards accurate generative models of video: A new metric \& challenges}.
\newblock \bibinfo{journal}{\emph{arXiv preprint arXiv:1812.01717}} (\bibinfo{year}{2018}).
\newblock


\bibitem[Vaswani(2017)]%
        {vaswani2017attention}
\bibfield{author}{\bibinfo{person}{A Vaswani}.} \bibinfo{year}{2017}\natexlab{}.
\newblock \showarticletitle{Attention is all you need}.
\newblock \bibinfo{journal}{\emph{Advances in Neural Information Processing Systems}} (\bibinfo{year}{2017}).
\newblock


\bibitem[Wang et~al\mbox{.}(2023b)]%
        {wang2023modelscope}
\bibfield{author}{\bibinfo{person}{Jiuniu Wang}, \bibinfo{person}{Hangjie Yuan}, \bibinfo{person}{Dayou Chen}, \bibinfo{person}{Yingya Zhang}, \bibinfo{person}{Xiang Wang}, {and} \bibinfo{person}{Shiwei Zhang}.} \bibinfo{year}{2023}\natexlab{b}.
\newblock \showarticletitle{Modelscope text-to-video technical report}.
\newblock \bibinfo{journal}{\emph{arXiv preprint arXiv:2308.06571}} (\bibinfo{year}{2023}).
\newblock


\bibitem[Wang et~al\mbox{.}(2023a)]%
        {wang2023disco}
\bibfield{author}{\bibinfo{person}{Tan Wang}, \bibinfo{person}{Linjie Li}, \bibinfo{person}{Kevin Lin}, \bibinfo{person}{Chung-Ching Lin}, \bibinfo{person}{Zhengyuan Yang}, \bibinfo{person}{Hanwang Zhang}, \bibinfo{person}{Zicheng Liu}, {and} \bibinfo{person}{Lijuan Wang}.} \bibinfo{year}{2023}\natexlab{a}.
\newblock \showarticletitle{Disco: Disentangled control for referring human dance generation in real world}.
\newblock \bibinfo{journal}{\emph{arXiv e-prints}} (\bibinfo{year}{2023}), \bibinfo{pages}{arXiv--2307}.
\newblock


\bibitem[Wang et~al\mbox{.}(2024a)]%
        {wang2024instancediffusion}
\bibfield{author}{\bibinfo{person}{Xudong Wang}, \bibinfo{person}{Trevor Darrell}, \bibinfo{person}{Sai~Saketh Rambhatla}, \bibinfo{person}{Rohit Girdhar}, {and} \bibinfo{person}{Ishan Misra}.} \bibinfo{year}{2024}\natexlab{a}.
\newblock \showarticletitle{Instancediffusion: Instance-level control for image generation}. In \bibinfo{booktitle}{\emph{Proceedings of the IEEE/CVF Conference on Computer Vision and Pattern Recognition}}. \bibinfo{pages}{6232--6242}.
\newblock


\bibitem[Wang et~al\mbox{.}(2019)]%
        {wang2019i3d}
\bibfield{author}{\bibinfo{person}{Xianyuan Wang}, \bibinfo{person}{Zhenjiang Miao}, \bibinfo{person}{Ruyi Zhang}, {and} \bibinfo{person}{Shanshan Hao}.} \bibinfo{year}{2019}\natexlab{}.
\newblock \showarticletitle{I3d-lstm: A new model for human action recognition}. In \bibinfo{booktitle}{\emph{IOP conference series: materials science and engineering}}, Vol.~\bibinfo{volume}{569}. IOP Publishing, \bibinfo{pages}{032035}.
\newblock


\bibitem[Wang et~al\mbox{.}(2024b)]%
        {wang2024videocomposer}
\bibfield{author}{\bibinfo{person}{Xiang Wang}, \bibinfo{person}{Hangjie Yuan}, \bibinfo{person}{Shiwei Zhang}, \bibinfo{person}{Dayou Chen}, \bibinfo{person}{Jiuniu Wang}, \bibinfo{person}{Yingya Zhang}, \bibinfo{person}{Yujun Shen}, \bibinfo{person}{Deli Zhao}, {and} \bibinfo{person}{Jingren Zhou}.} \bibinfo{year}{2024}\natexlab{b}.
\newblock \showarticletitle{Videocomposer: Compositional video synthesis with motion controllability}.
\newblock \bibinfo{journal}{\emph{Advances in Neural Information Processing Systems}}  \bibinfo{volume}{36} (\bibinfo{year}{2024}).
\newblock


\bibitem[Wei et~al\mbox{.}(2024)]%
        {wei2024aniportrait}
\bibfield{author}{\bibinfo{person}{Huawei Wei}, \bibinfo{person}{Zejun Yang}, {and} \bibinfo{person}{Zhisheng Wang}.} \bibinfo{year}{2024}\natexlab{}.
\newblock \showarticletitle{Aniportrait: Audio-driven synthesis of photorealistic portrait animation}.
\newblock \bibinfo{journal}{\emph{arXiv preprint arXiv:2403.17694}} (\bibinfo{year}{2024}).
\newblock


\bibitem[Xiao et~al\mbox{.}(2022)]%
        {xiaotackling}
\bibfield{author}{\bibinfo{person}{Zhisheng Xiao}, \bibinfo{person}{Karsten Kreis}, {and} \bibinfo{person}{Arash Vahdat}.} \bibinfo{year}{2022}\natexlab{}.
\newblock \showarticletitle{Tackling the Generative Learning Trilemma with Denoising Diffusion GANs}. In \bibinfo{booktitle}{\emph{International Conference on Learning Representations}}.
\newblock


\bibitem[Xiong et~al\mbox{.}(2024)]%
        {xiong2024segtalker}
\bibfield{author}{\bibinfo{person}{Lingyu Xiong}, \bibinfo{person}{Xize Cheng}, \bibinfo{person}{Jintao Tan}, \bibinfo{person}{Xianjia Wu}, \bibinfo{person}{Xiandong Li}, \bibinfo{person}{Lei Zhu}, \bibinfo{person}{Fei Ma}, \bibinfo{person}{Minglei Li}, \bibinfo{person}{Huang Xu}, {and} \bibinfo{person}{Zhihui Hu}.} \bibinfo{year}{2024}\natexlab{}.
\newblock \showarticletitle{SegTalker: Segmentation-based Talking Face Generation with Mask-guided Local Editing}. In \bibinfo{booktitle}{\emph{Proceedings of the 32nd ACM International Conference on Multimedia}}. \bibinfo{pages}{3170--3179}.
\newblock


\bibitem[Xu et~al\mbox{.}(2024b)]%
        {xu2024hallo}
\bibfield{author}{\bibinfo{person}{Mingwang Xu}, \bibinfo{person}{Hui Li}, \bibinfo{person}{Qingkun Su}, \bibinfo{person}{Hanlin Shang}, \bibinfo{person}{Liwei Zhang}, \bibinfo{person}{Ce Liu}, \bibinfo{person}{Jingdong Wang}, \bibinfo{person}{Yao Yao}, {and} \bibinfo{person}{Siyu Zhu}.} \bibinfo{year}{2024}\natexlab{b}.
\newblock \showarticletitle{Hallo: Hierarchical audio-driven visual synthesis for portrait image animation}.
\newblock \bibinfo{journal}{\emph{arXiv preprint arXiv:2406.08801}} (\bibinfo{year}{2024}).
\newblock


\bibitem[Xu et~al\mbox{.}(2024a)]%
        {xu2024vasa}
\bibfield{author}{\bibinfo{person}{Sicheng Xu}, \bibinfo{person}{Guojun Chen}, \bibinfo{person}{Yu-Xiao Guo}, \bibinfo{person}{Jiaolong Yang}, \bibinfo{person}{Chong Li}, \bibinfo{person}{Zhenyu Zang}, \bibinfo{person}{Yizhong Zhang}, \bibinfo{person}{Xin Tong}, {and} \bibinfo{person}{Baining Guo}.} \bibinfo{year}{2024}\natexlab{a}.
\newblock \showarticletitle{Vasa-1: Lifelike audio-driven talking faces generated in real time}.
\newblock \bibinfo{journal}{\emph{arXiv preprint arXiv:2404.10667}} (\bibinfo{year}{2024}).
\newblock


\bibitem[Yang et~al\mbox{.}(2024)]%
        {yang2024consistentavatar}
\bibfield{author}{\bibinfo{person}{Haijie Yang}, \bibinfo{person}{Zhenyu Zhang}, \bibinfo{person}{Hao Tang}, \bibinfo{person}{Jianjun Qian}, {and} \bibinfo{person}{Jian Yang}.} \bibinfo{year}{2024}\natexlab{}.
\newblock \showarticletitle{ConsistentAvatar: Learning to Diffuse Fully Consistent Talking Head Avatar with Temporal Guidance}. In \bibinfo{booktitle}{\emph{Proceedings of the 32nd ACM International Conference on Multimedia}}. \bibinfo{pages}{3964--3973}.
\newblock


\bibitem[Yang et~al\mbox{.}(2023)]%
        {yang2023effective}
\bibfield{author}{\bibinfo{person}{Zhendong Yang}, \bibinfo{person}{Ailing Zeng}, \bibinfo{person}{Chun Yuan}, {and} \bibinfo{person}{Yu Li}.} \bibinfo{year}{2023}\natexlab{}.
\newblock \showarticletitle{Effective whole-body pose estimation with two-stages distillation}. In \bibinfo{booktitle}{\emph{Proceedings of the IEEE/CVF International Conference on Computer Vision}}. \bibinfo{pages}{4210--4220}.
\newblock


\bibitem[Yao et~al\mbox{.}(2024)]%
        {yao2024fd2talk}
\bibfield{author}{\bibinfo{person}{Ziyu Yao}, \bibinfo{person}{Xuxin Cheng}, {and} \bibinfo{person}{Zhiqi Huang}.} \bibinfo{year}{2024}\natexlab{}.
\newblock \showarticletitle{FD2Talk: Towards Generalized Talking Head Generation with Facial Decoupled Diffusion Model}. In \bibinfo{booktitle}{\emph{Proceedings of the 32nd ACM International Conference on Multimedia}}. \bibinfo{pages}{3411--3420}.
\newblock


\bibitem[Yoon et~al\mbox{.}(2020)]%
        {yoon2020speech}
\bibfield{author}{\bibinfo{person}{Youngwoo Yoon}, \bibinfo{person}{Bok Cha}, \bibinfo{person}{Joo-Haeng Lee}, \bibinfo{person}{Minsu Jang}, \bibinfo{person}{Jaeyeon Lee}, \bibinfo{person}{Jaehong Kim}, {and} \bibinfo{person}{Geehyuk Lee}.} \bibinfo{year}{2020}\natexlab{}.
\newblock \showarticletitle{Speech gesture generation from the trimodal context of text, audio, and speaker identity}.
\newblock \bibinfo{journal}{\emph{ACM Transactions on Graphics (TOG)}} \bibinfo{volume}{39}, \bibinfo{number}{6} (\bibinfo{year}{2020}), \bibinfo{pages}{1--16}.
\newblock


\bibitem[Zhang et~al\mbox{.}(2024a)]%
        {zhang2024personatalk}
\bibfield{author}{\bibinfo{person}{Longhao Zhang}, \bibinfo{person}{Shuang Liang}, \bibinfo{person}{Zhipeng Ge}, {and} \bibinfo{person}{Tianshu Hu}.} \bibinfo{year}{2024}\natexlab{a}.
\newblock \showarticletitle{Personatalk: Bring attention to your persona in visual dubbing}. In \bibinfo{booktitle}{\emph{SIGGRAPH Asia 2024 Conference Papers}}. \bibinfo{pages}{1--9}.
\newblock


\bibitem[Zhang et~al\mbox{.}(2023)]%
        {zhang2023adding}
\bibfield{author}{\bibinfo{person}{Lvmin Zhang}, \bibinfo{person}{Anyi Rao}, {and} \bibinfo{person}{Maneesh Agrawala}.} \bibinfo{year}{2023}\natexlab{}.
\newblock \showarticletitle{Adding conditional control to text-to-image diffusion models}. In \bibinfo{booktitle}{\emph{Proceedings of the IEEE/CVF International Conference on Computer Vision}}. \bibinfo{pages}{3836--3847}.
\newblock


\bibitem[Zhang et~al\mbox{.}(2024b)]%
        {zhang2024tora}
\bibfield{author}{\bibinfo{person}{Zhenghao Zhang}, \bibinfo{person}{Junchao Liao}, \bibinfo{person}{Menghao Li}, \bibinfo{person}{Zuozhuo Dai}, \bibinfo{person}{Bingxue Qiu}, \bibinfo{person}{Siyu Zhu}, \bibinfo{person}{Long Qin}, {and} \bibinfo{person}{Weizhi Wang}.} \bibinfo{year}{2024}\natexlab{b}.
\newblock \showarticletitle{Tora: Trajectory-oriented diffusion transformer for video generation}.
\newblock \bibinfo{journal}{\emph{arXiv preprint arXiv:2407.21705}} (\bibinfo{year}{2024}).
\newblock


\bibitem[Zhou et~al\mbox{.}(2021)]%
        {zhou2021pose}
\bibfield{author}{\bibinfo{person}{Hang Zhou}, \bibinfo{person}{Yasheng Sun}, \bibinfo{person}{Wayne Wu}, \bibinfo{person}{Chen~Change Loy}, \bibinfo{person}{Xiaogang Wang}, {and} \bibinfo{person}{Ziwei Liu}.} \bibinfo{year}{2021}\natexlab{}.
\newblock \showarticletitle{Pose-controllable talking face generation by implicitly modularized audio-visual representation}. In \bibinfo{booktitle}{\emph{Proceedings of the IEEE/CVF conference on computer vision and pattern recognition}}. \bibinfo{pages}{4176--4186}.
\newblock


\bibitem[Zhu et~al\mbox{.}(2025)]%
        {zhu2025champ}
\bibfield{author}{\bibinfo{person}{Shenhao Zhu}, \bibinfo{person}{Junming~Leo Chen}, \bibinfo{person}{Zuozhuo Dai}, \bibinfo{person}{Zilong Dong}, \bibinfo{person}{Yinghui Xu}, \bibinfo{person}{Xun Cao}, \bibinfo{person}{Yao Yao}, \bibinfo{person}{Hao Zhu}, {and} \bibinfo{person}{Siyu Zhu}.} \bibinfo{year}{2025}\natexlab{}.
\newblock \showarticletitle{Champ: Controllable and consistent human image animation with 3d parametric guidance}. In \bibinfo{booktitle}{\emph{European Conference on Computer Vision}}. Springer, \bibinfo{pages}{145--162}.
\newblock


\end{thebibliography}
